# Knowledge-based Biomedical Data Science 2019

Tiffany J. Callahan[1], Harrison Pielke-Lombardo[1], Ignacio J. Tripodi[1,2], Lawrence E. Hunter[1]*

[1]Computational Bioscience Program, Department of Pharmacology, University of Colorado Denver Anschutz Medical Campus, Aurora, CO USA

[2]Computer Science, University of Colorado Boulder, Boulder, CO USA

***Corresponding Author:**

Email: larry.hunter@cuanschutz.edu
Address: Computational Bioscience Program, University of Colorado-Denver Anschutz Medical Campus, Mail Stop 8303, P18-6129, 12800 E. 19th Avenue, Aurora, CO 80045

**ORCIDs:**

Callahan: https://orcid.org/0000-0002-8169-9049
Pielke-Lombardo: https://orcid.org/0000-0002-8865-6025
Tripodi: https://orcid.org/0000-0003-3066-4273
Hunter: https://orcid.org/0000-0003-1455-3370

## ABSTRACT

Knowledge-based biomedical data science (KBDS) involves the design and implementation of computer systems that *act as if they knew* about biomedicine. Such systems depend on formally represented knowledge in computer systems, often in the form of knowledge graphs. Here we survey the progress in the last year in systems that use formally represented knowledge to address data science problems in both clinical and biological domains, as well as on approaches for creating knowledge graphs. Major themes include the relationships between knowledge graphs and machine learning, the use of natural language processing, and the expansion of knowledge-based approaches to novel domains, such as Chinese Traditional Medicine and biodiversity.

## INTRODUCTION

### What is Knowledge-based Biomedical Data Science?

Knowledge-based biomedical data science (KBDS) involves the design and implementation of computer systems that *act as if they knew* about biomedicine.[1] There are many ways in which a system might act as if it knew something: for example, it might be able to use existing knowledge to generate, rank or evaluate hypotheses about a dataset, or able to answer a natural language question about a biomedical topic.

Knowledge-based systems have long been a theme in artificial intelligence research. Knowledge-based systems specify a *knowledge representation* -- how a computer system represents knowledge internally -- and one or more methods of *inference or reasoning* -- how computations over those representations (perhaps combined with other inputs) are used to produce outputs. Classical descriptions of knowledge representation and reasoning systems, e.g. (2) characterize them by the ontological commitments a knowledge representation makes (that is, what it can or cannot describe), which inferences are possible within it, and, sometimes, which of those inferences can be made efficiently. These issues remain useful in thinking about how knowledge representation and reasoning play a role in today's data science environment.

### Ontologies

Knowledge representations are said to be *grounded* in a set of primitive terms that specify those ontological commitments: the entities and processes that can be referred to by that knowledge representation. Computational ontologies are collections of primitives relevant to a domain, often related to each other by explicit subsumption (subclass of) and meronomy (part of) statements. Biomedical ontologies (such as the Gene Ontology (3)) are community consensus views of the entities involved in biology, medicine and biomedical research, analogous to how nomenclature committees systematize naming conventions. knowledge-bases created using primitives from community-curated ontologies, rather than using idiosyncratic or single-use sets of primitives, provides significant advantages for reproducibility in scientific research, for interoperability, and in avoiding pitfalls in the modeling of knowledge. While lacking some useful aspects of a computational ontology, terminological resources such as UMLS (4), SNOMED CT (5) and the NCI thesaurus (6) have also been used to provide interoperable foundations for knowledge representations.

[1] This is a revised and extended version of the introduction to (1)

Primitives can be combined into *assertions* that express facts about the world. In the simplest case, an assertion links two represented elements with a specific relationship. Consider, for example, that the Protein Ontology contains a representation of *human TP53*, the Gene Ontology contains a representation of the process of *DNA strand renaturation*, and the Relation Ontology contains a representation of *participation* that can link a physical entity to a process in which it participates. Those three ontological entities can be composed into an assertion that could be part of a knowledge-base: *human TP53 participates in DNA strand renaturation*. Collections of assertions, generally called knowledge-bases, can be created and shared, and then in turn used by other systems that apply various inference methods to fulfill particular application needs.

**Standards**

While ontologies provide the primitive elements from which a knowledge representation is constructed, they are agnostic about the mechanisms by which entities are assembled into assertions. In 2011, the World Wide Web Consortium promulgated a collection of international standards for linking entities with shared meaning into assertions and managing collections of assertions, together referred to as the Semantic Web (7). The Semantic Web builds on the standard Resource Description Framework (RDF) (8), which provides a way to link three uniform resource identifiers (URIs) (9) to specify a relationship between a pair of entities (forming an RDF "triple"). Collections of triples form a *graph*, where the entities are nodes and the relationships are edges connecting them. A computational mechanism for managing such collections is called a triple store (10). The Semantic Web standards also define RDF Schemas (RDFS) and a Web Ontology Language (OWL) which provide additional expressivity, SPARQL (SPARQL Protocol and RDF Query Language), which provides a query language for interrogating RDF graphs or triple stores, and the Simple Knowledge Organization System, which provides a basic ontology.

The web ontology language (OWL) (11) specifies two types of entities: instances and classes. Instances are particular entities or processes in the world (e.g., a particular molecule of TP53) and classes are groupings of instances that meet a defined set of individually necessary and collectively sufficient criteria (e.g. human TP53 proteins). As it lacks variables and quantification, OWL cannot express all logical statements about primitives; the subset of first order logic that OWL can express is inspired by description logics. For reasons rooted in description logic theory, in OWL ontologies are called

*TBoxes* (for terminology), and assertions composed of them are *ABoxes* (for assertion) (12).

### Knowledge-base vs. Knowledge Graph

Knowledge-bases that can be represented as graphs are often called *knowledge graphs*. While not all knowledge-bases are implemented as graphs (e.g. some are databases where table structure makes implicit assertions), in recent years, it has become very common to represent knowledge-bases using the Semantic Web standard or, at least be able to produce and consume Semantic Web compatible versions. For that reason, the terms knowledge-base and knowledge graph are often used interchangeably. In 2012, Google announced its proprietary Knowledge Graph, which also popularized the use of the term (13). The literature sometimes contains terminological imprecision about what the differences are between knowledge-bases, knowledge graphs and ontologies; there is a review and analysis of various published definitions (14). In this review, we use the term *knowledge graph* (or KG) and say a KG is *grounded* in the set of primitives from which it is constructed. Some KGs also include a set of logical rules that relate assertions to each other (e.g. Human TP53 is the subclass of TP53 proteins that is found in the organism human) called *axioms*.

### Biomedical Applications

KBDS does computation over KGs (and perhaps other inputs) to make inferences about biomedicine. While each of the publications surveyed below addresses different problems using different techniques, there are some common themes in the computational approaches to using KGs.

A major use of KGs is simply to organize knowledge for information retrieval. Such systems are designed to make it possible to find facts or evidence regarding a wide variety of topics, ranging in this review from cataloging traditional Chinese medical practices to decision support for pharmacovigilance. KGs have also been used to improve other forms of information retrieval, such as finding relevant publications in the literature.

Computer science has produced many algorithms that operate on graphs, and therefore on KGs. One particular class of graph algorithm that is widely used in KBDS is *edge (*or *link) prediction.* Edge prediction methods generally use the structure of a graph to identify edges that are likely but missing in the graph. In KGs, these are predictions of unrepresented facts about the world. This is a form of hypothesis generation, and often includes an estimate of the confidence in the prediction. Many approaches to drug repurposing in this review use edge prediction algorithms over KGs of drugs and diseases to identify new indications. Another broad class of graph algorithms does *community finding*, or identification of groups of entities in a KG that are similar or highly related to each other. For example, some approaches to disease sub-phenotyping apply community-finding approaches to KGs encoding information about patients.

Machine learning, particularly in the form of artificial neural networks, is also widely used in the KG context. One frequent application of neural networks to KGs is to create vector embeddings of entities or assertions by training auto-encoder networks with inputs constructed from the KG. These embeddings can then in turn be used to compute knowledge-based similarities, e.g. between drugs, proteins, and diseases. Neural network methods have also been used to identify parts of a KG relevant to answering an input question.

The Semantic Web OWL standard was designed to facilitate two important classes of reasoning over KGs: *satisfiability* and *subsumption inference.* It is possible for a KG to define a class that has no members (e.g. TP53 homologs in bacteria); satisfiability inference checks to see if a class definition is *logically* satisfiable. Subsumption inference uses class definitions to identify all classes that are fully contained within some other class (e.g. all proteins are nitrogen-containing compounds). Specific reasoners, such as ELK (15) or Hermit (16) can be used to make these inferences with particular computational performance guarantees, which can be important in large KGs. Subsumption inference in particular is useful in KBDS because it makes explicit many edges that are otherwise implicit in KGs, and therefore can improve the results of other algorithms that depend on the structure of the graph, such as link prediction or vector embeddings.

**Known Challenges**

Computational performance is a challenge in other areas as well. Biomedical knowledge is very extensive, and broad biomedical KGs can contain billions of assertions. A wide variety of schemes have been proposed to address the computational complexity of both querying and inference over KGs (17).

A few KGs (e.g. Gene Ontology annotations (18), GO-CAMs (19), or Reactome (20)) are constructed through painstaking and expensive manual curation efforts. However, several algorithmic approaches have been proposed to either augment these efforts or fully automate them. Automated approaches to KG construction fall into two broad classes: natural language processing (NLP) and data-driven. Data-driven KG construction can involve the integration of previously disparate resources, or the direct analysis of large-scale datasets.

NLP methods propose to extract information from a set of documents to create a KG, (e.g. SemMedDB (21)). As NLP methods are all imperfect, these approaches are often focused on assessing the reliability of the information extracted, or on techniques to manage missing or erroneous assertions and other sources of noise.

Some data-driven approaches simply transform existing databases (e.g. DrugBank (22)) into KG form, which can facilitate adherence to FAIR (Findable, Accessible, Interoperable, and Reusable) research principles (23). More frequently, data-driven KG construction integrates of multiple sources of data into a single KG. If an integrated KG can ground the different sources to one set of primitives (ideally from a community-curated ontology), that facilitates inference over the combined information. As there are thousands of public, biomedically important databases (24), integration approaches that support semantic compatibility are important. Integration also can lead to improved data quality, as incompatibilities sometimes signal errors (25).

## METHODS

### Review Methodology

Relevant literature was obtained by searching PubMed and Google Scholar using the following phrases: '"*knowledge graph", "biomedicine*"', '"*knowledge graph", "medicine*"', '"*knowledge graph", "medical*"', and '"*knowledge graph", "biology*"'. The same set of terms were also searched by replacing "graph" with "base". All article types were eligible for inclusion (i.e. conference proceedings, dissertations, book chapters, pre-reviewed archived manuscripts, and published peer-reviewed manuscripts).

## RESULTS

The search phrases above returned 52 papers from PubMed and 7,752 papers from Google Scholar. Manual review of these papers was performed to identify those that were focused on the use or construction of KGs within the biomedical domain resulted, which resulted in a reduced set of 174 papers. This set of papers was then further reduced to only include papers published or posted to public manuscript archives within last year (January 2018-September 2019) whose full-text version was publicly available at the time of review, resulting in a final set of 83 papers reviewed here. The final set of papers was further broken down by year of publication, the publication type, and the publication venue (i.e. the journal or archive name). Among the 83 papers, 44 were published in 2018. The majority of papers were published in conference proceedings (n=39) or in peer-reviewed journals (n=25), with the remaining papers published as online preprints (n=19). Among these, the majority of the 2018 papers were submitted to peer-reviewed journals (64.0%), whereas most of the 2019 papers were submitted to arxiv (73.7%). The number of conference submissions increased slightly between 2018 and 2019 (56.4%); 2018 papers were primarily submitted to the Institute of Electrical and Electronics Engineers (n=11) whereas 2019 papers were submitted to the Association for Computing Machinery (n=4) and the Association for the Advancement of Artificial Intelligence (n=2).

Information about each paper included in the final set is presented in Tables 1 and 2 (with more detailed tables in supplementary materials), and broad themes spanning multiple papers are described below.

## Organization and Presentation of Findings

These publications fall into two broad categories, which we use to organize the review: *application* of KGs (n=53) versus *production* of KGs (n=30). Applications are noted in a wide variety of biomedical research domains, ranging from analysis of genomic data to clinical decision support. There is also a close relationship between KGs and biomedical NLP: KGs can be used to improve the quality of NLP, and NLP can be used to generate KGs from the literature. We conclude by considering some nascent projects likely to be important in the near future, characterizing current barriers to building and using biomedical KGs, and making some recommendations.

## Applications of Knowledge Graphs in Biomedical Data Science

Table 1 provides a high-level summary of the reviewed papers which applied KGs to help solve a biomedical data science problem.

### Clinical Applications

There were three primary themes identified within this domain, including the use of KGs to improve the retrieval of information from the literature or from large sources of clinical data (26–30); the use of KGs to provide confidence either by adding evidence to support phenomena observed in data (31–35) or by completing missing information and deriving new hypotheses (36–40); and the use of KGs to improve the representation of complicated patient data or personal health information (41, 42) or presentation of complex information or results (43–45).

KGs can be used to refine user queries and otherwise improve information retrieval from the literature or from an EHR system. One study demonstrated using KGs with traditional rule-based approaches for information retrieval performed better than using either approach alone (26). Liu et al. (27) proposed a novel graph-based representation of patient data where entities were linked to concepts in a biomedical KG in order to enable querying based on domain knowledge. Other clinical applications of KGs in information retrieval included a KG-based component added to a larger system improved the ability of doctors to identify meaningful information from an EHR (28), a KG-based method for users to formulate queries within the context of relevant domain knowledge (29), and a system to re-write user queries using domain knowledge (30).

An important application of KGs is to address clinical uncertainties by identifying relevant evidence. KGs have been leveraged to provide evidence for diagnostic assistance or clinical decision (33) support machinery or surveillance. For example, Bakal et al, used SemMedDB and a subset of the UMLS to better predict treatments for and causes of different diseases (31) and Reumann et al. (32) found that using a KG was helpful for correctly identifying rare disease patients when examined using over 100 different queries (32). There were two articles which focused on surveillance: Bobed et al. (34) built a KG from an ontology (OntoADR) and a clinical terminology (SNOMED CT) as means to improve pharmacovigilance and Kamdar et al. (35) built a KG using drug classes from the Anatomical Therapeutic Chemical Classification System and active ingredients in RxNorm to better understand opioid use patterns across the United States.

Link prediction approaches over KGs were used to discover missing knowledge or generate hypotheses, e.g. in (27). To improve the identification of comorbid diseases, Biswas, Mitra & Rao (37) built a KG using the approach outlined by (36) and then performed link prediction using an inductive inference method. Also using inductive inference methods, Callahan et al. (38) described a method for transforming OWL-encoded knowledge to create representations that were better suited to inductive inference tasks; the results were evaluated using queries against KaBOB (46). Neil et al. (39) described a method for transforming KGs into graph convolutional neural networks and an attention model using independent learnable weights to measure of each edge's "usefulness".

KGs are also used to capture complex patient information for further processing. Xie et al. (41) created patient-specific traditional Chinese medicine (TCM) KGs by mapping patient data to a general TCM-specific KG. Shang et al. (42) describes a method for creating visit-level representations of patients from EHR data and mapping to a drug-drug interaction KG to provide personalized medication combination recommendations. Three articles focused on how to improve the presentations of complex information or results. Huang et al. (43) developed a novel tool to enrich and visualize patient data by incorporating KG embeddings. Singh et al. (44) develop an interactive tool built on Cytoscape (47) to help users interact with their network data (44). Queralt-Rosinach et al. (45) introduced a novel approach to create custom systematic literature reviews by formulating the review as a biomedical KG that contains information relevant to specific hypotheses provided by a user.

## Biological Applications

In more basic research applications, broad themes included the use of KGs to produce vector embeddings for prediction or visualization in low dimensional spaces (17, 40, 48); the use of link prediction methods over KGs to hypothesize previously unobserved relationships (38, 40, 42, 49–58); and the use of KGs to generate complex mechanistic accounts of experimental data . Several efforts combined these themes, particularly the use of edge embeddings to improve link prediction (37, 50, 55, 59, 60).

Node and edge embeddings provide a powerful method to suggest relationships among entities via similarity functions, in ways that complement path traversal through the graph. Semantic similarity inspired hypotheses are valuable include drug-drug (42, 49, 53), drug-target (53), or protein-protein interactions (48, 50), many of which were in turn applied to drug repurposing. KG-based embeddings into low dimensional spaces were also used to visualize clusters in two- or three-dimensional projections (43) to better display entities of interest.

In a particularly innovative approach, Tripodi et al. (40) combined gene expression time series and KG embeddings from a Human-centric KG (61) to create specific and detailed hypotheses regarding mechanisms of toxicity. The KG subgraphs that made up the hypothesized mechanisms were far richer than black-box toxicity predictions and were also used to generate natural language narratives describing the mechanisms and the evidence for them.

## Natural Language Processing Applications

KGs have been used to improve NLP performance in a wide variety of genres, including summarization or information extraction from EHRs and answering medical questions (17, 28, 29, 33, 42, 62, 63). KG-derived embeddings used alone, or in combination with text-derived features (48) improved performance of a variety of NLP tasks, including named-entity recognition (64), coreference resolution (65) and relation extraction (66). Several applications demonstrate the utility of KGs in information extraction methods. Ontologies can serve as formal dictionaries allowing for rapid indexing in named entity recognition and word-sense disambiguation tasks (67, 68). Compared to lexicons, KGs offer far richer semantic context, identifying not only similar concepts, but a rich collections of relationships that can be used to disambiguate or otherwise improve concept recognition in texts (67–70).

## **Constructing Knowledge Graphs**

As the applications of KGs are many and varied, and their construction is demanding, substantial effort was reported in methods and new results in the construction of KGs, as well as extending, integrating and evaluating them. Table 2 provides a high-level summary of these publications.

Last year saw the announcement of a new approach to Gene Ontology (GO) annotation called Causal Activity Models, or GO-CAMs (19). Although GO annotations are perhaps the most widely used knowledge representations in biomedical research, until GO-CAMs were introduced, the annotations could not be assembled into a coherent knowledge graph. While individual annotations implicitly linked GO classes to gene products, contextual information was lost -- for example, the annotation process could not capture that cytochrome C participated in apoptosis only when it was in the cytoplasm. GO-CAM models, and associated tooling is gradually replacing the traditional GO annotation process within the Alliance for Genomic Resources, meaning future GO annotation will produce an increasingly rich, manually curated KG.

Other efforts to produce domain-specific KGs were published in a variety of areas, including biodiversity (71–73), the microbiome (74), and for the purpose of enriching clinical data (75, 76). The articles on biodiversity focused specifically on how a KG could be created and linked to identifiers in the literature (71, 72) or other important biodiversity resources (73). In contrast, papers using KGs for clinical enrichment aimed to use KGs as way to link clinical data to sources of evidence to provide support to clinical observations (76–78) or to help make the data more interpretable with respect to underlying biological mechanism(s) (75) for improved diagnosis (79).

Historically, NLP information extraction efforts have often been used to construct KGs; two novel methods to do so were published last year. One proposed a minimum supervision-based approach which combined a traditional NLP-pipelines for information extraction with the use of biomedical context embeddings (80). The other focused on improving the extraction of biomedical facts from the literature by leveraging and refining specific seed patterns (81).

Although not a construction method *per se,* (82) presented an evaluation of one of the most widely used NLP-constructed KGs, SemMedDB. A large number of contradictory assertions were found in a variety of fundamental relationship categories, underscoring the need to be cautious regarding noise in NLP-derived KGs.

Finally, an ontology called BioKNO and a set of associated tools leveraging OWL was presented to assist scientists attempting to share data according FAIR principles (83).

**Organizational efforts in knowledge-based biomedical data science**

Both US and European scientific institutions support KG efforts. Perhaps the most ambitious of these is the National Institutes of Health's National Center for Advancing Translational Science's Biomedical Data Translator project (84). The goal of the Translator is a computational system that integrates sources of existing biomedical knowledge in order to translate clinical inquiries into relevant biomedical research results which synthesize elements of the integrated knowledge to directly answer the inquiry or generate testable hypotheses (85). A recent funding call[2] targets $13.5 million per year for up to five years towards the construction of "Knowledge Providers" and "Autonomous Relay Agents." Knowledge providers are systems that seek out, integrate and provide high-value data sources within a specific scope of Translator-relevant knowledge, and presumably would primarily use KGs to do so. Relay agents are to take clinical queries in a standardized format, dispatch subtasks to appropriate knowledge providers, receive responses back from knowledge sources (presumably also as subgraphs of a KG), and process responses using scoring metrics in order to return the most relevant and highest quality potential responses.

Elixir Europe (86) is a large multinational (and European Commission) project with the goal of managing and safeguarding the data generated by publicly funded life science research and integrating bioinformatics resources. In pursuit of those goals, Elixir's

2 https://ncats.nih.gov/files/NCATS-Translator-FY20-COMBINED-FOA-FINAL.pdf

interoperability platform promotes efforts in the European life science community to adopt standardized file formats, metadata, vocabulary and identifiers, including efforts in Semantic Web and adoption of community-curated ontologies. The Elixir Core Data Resources (87) are leaders in the production of interoperable knowledge resources, and are widely used components of biomedical KGs.

## CONCLUSIONS AND RECOMMENDATIONS

### Recommendations

Knowledge Graph Embeddings

A very active area of research is using KGs to create knowledge-based vector embeddings. Good vector embeddings are important to the performance of machine learning systems, and therefore have wide applicability. KGs have been used to create embeddings for entities of many kinds (ranging from genes to patients), as well as for relations, assertions, and more complex representations. Applications of these embeddings include prediction of drug-drug interactions, drug-target interactions, target discovery, finding clinically relevant evidence and more. In addition to being able to re-use embeddings from the surveyed papers, we recommend considering the tools described in the BioKEEN paper that describes a Python-based library for training and tuning models to produce new knowledge-based embeddings.

Natural Language Processing-based Knowledge Graphs

A major theme in the literature is the use of text mining and natural language processing techniques to generate KGs. While this approach offers the potential for breadth missing from most manually curated KGs, it comes at the cost of a large proportion of errors. The Cong et al. (82), paper evaluated SemMedDB, a widely used KG produced by the US National Library of Medicine, and found nearly half a million inconsistent assertions, as well as a wide variety of apparently missing relationships . While they suggested methods that could be used to improve the quality of SemMedDB, our recommendation is to recognize that NLP-generated KGs are likely to be very noisy, and need to be used with caution.

## Barriers and Future Work

### Barriers

Current barriers to constructing and using KGs includes navigating everything from KG and data availability, data licensing issues (sometimes there are different licenses for each data source), a lack of agreed upon standards for constructing KGs to dependency upon resources (i.e. software languages or applications) which may be obsolete, deprecated, or outside of the users skill set or area of expertise.

The construction of KGs is a difficult task, so re-use of existing resources is desirable whenever possible. However, there are challenges in applying each of the existing KGs (Table 3) to new tasks. While GO-CAMs have great promise, as of this writing, only a relatively small number of them have been curated. Reactome provides a very high quality and extensive KG grounded in community-curated ontologies, but it is limited in scope to biochemical reactions and pathways. Other manually annotated resources (such as GO annotations) cannot be straightforwardly assembled into KGs amenable to OWL reasoners. Data integration derived KGs such as KaBOB are both broad and grounded in community curated ontologies, but licensing restrictions mean that users have to download software and build the KG themselves, which requires expertise and computational resources. Hetio-nets are broad, but not grounded in community-curated ontologies, which makes integrating it with other resources difficult. Bio2RDF is a mashup of many different kinds of data without a consistent set of primitives, allowing inconsistencies in the represented knowledge. Finally, NLP derived systems such as SemMedDB are noisy, making trustworthiness an issue.

Automatic KG construction from literature sources is usually framed as a relation extraction problem, where semantic triples are inferred from text, and then assembled into a KG. The “correctness” of this approach to KG construction can be evaluated either before the KG is constructed, which involves evaluating the relation extraction process itself (a more traditional approach), or by evaluating the quality of the resulting KG itself. Evaluating the quality of the constructed KG allows for the use of the reasoning methods described above.

The lack of standards for constructing KGs within the biomedical domain may be one of the reasons why they are challenging to evaluate. Of the reviewed articles on constructing KGs that evaluated their KGs (n=26), four provided qualitative evaluation (e.g., case studies or domain expert review of results, conceptual models, or prototypes, and focus groups), five provided quantitative evaluation (e.g., most often by applying a machine learning model to a specific set of held out data or to a new dataset or by performing a KG completion task like edge prediction), and 17 provided both types of evaluation. Of note, one of the articles that provided both types of evaluation utilized crowdsourcing as a means to validate triples from their KG (78).

## Future Work

There trends we observed in last year's work are likely to continue. Applications of KGs will likely continue to involve generations of embeddings and other uses of KGs in machine learning. The close relationships between development of NLP methods and KGs is likely to persist. The expansion of KG to areas beyond molecular biology (e.g., Biodiversity and Chinese Traditional Medicine this year) is also likely to continue. Some previous areas of research, e.g. KG-based enrichment analysis for gene sets, that did not see new results this year may also continue to be fruitful.

New methods applying KGs to analyze different sorts of experimental data (e.g. images) seem ripe for development. Robust and biologically meaningful ways to incorporate or add experimental data to biomedical KGs would help to improve the precision of predictions when used to generate novel hypotheses or a means for helping to interpret experimental results. Similarly, for clinical KGs, it will be important to find clinically meaningful ways to incorporate quantitative measures (e.g. laboratory test results and biomarker measurements) and outcomes from EHR data.

Other than the tables in this review, there exists no central reference site or repository one can access to identify all available biomedical KGs. A more systematic approach to sharing, documenting and encouraging interoperability among biomedical KGs. Existing efforts on general frameworks and tools like BioKEEN (88), Protégé (89), BlazeGraph (90), as well as all of the teams who maintain OBOs (91) are important, and could be extended towards further standardized tool development. As KG evaluation remains challenging, new methods or benchmarks would be valuable there.

A final area for hope for future work is on general tools for interacting interaction with KGs. SPARQL's limits, e.g. in pathway search (92) and ease of use make broader adoption challenging.

## DISCLOSURE STATEMENT

The authors have no conflicts of interest to disclose.

## ACKNOWLEDGEMENTS

The authors have no acknowledgements.

## SUPPLEMENTAL MATERIAL

The full versions of Tables 1-3 are submitted as Supplemental Tables 1-3. The supplemental material has been submitted in a separate word document titled: Supplementtal_Material.docx.

## Terms and Definitions List

Relevant terms and definitions from the manuscript.

| Abbreviation | Definition |
|---|---|
| DNA | Deoxyribonucleic Acid |
| EHR | Electronic Health Record |
| FAIR | Findable, Accessible, Interoperable, and Reusable |
| GO-CAM | Gene Ontology Causal Activity Modeling |
| KBDS | Knowledge-Based Biomedical data Science |
| KG | Knowledge Representation |
| NCI | National Cancer Institute |
| NLP | Natural Language Processing |
| Q&A | Question and Answer |
| OBO | Open Biomedical Ontologies |
| OWL | Web Ontology Language |
| RDF | Resource Description Framework |
| RDFS | Resource Description Framework Schemas |
| REST | Representational State Transfer |
| SemMedDB | Semantic MEDLINE Database |
| SKOS | Simple Knowledge Organization System |
| SPARQL | SPARQL Protocol and RDF Query Language |

## TABLES

Table 1 Reviewed Articles on Applications of KGs in Biomedical Data Science

| Article | Objective[a] | Evaluation Type | Availability | |
|---|---|---|---|---|
| | | | Code | UI |
| Exploiting Semantic Patterns Over Biomedical Knowledge Graphs For Predicting Treatment And Causative Relations (31) | Build high accuracy supervised predictive models to predict previously unknown treatment and causative relations between biomedical entities based only on semantic graph pattern features extracted from biomedical knowledge graphs | Both | TRUE | TRUE |
| Retrieval Method Of Electronic Medical Records Based On Rules And Knowledge Graph (26) | Demonstrate how knowledge graphs improve the retrieval of information from electronic health records | Quantitative | NA | NA |
| PrTransH: Embedding Probabilistic Medical Knowledge From Real World EHR Data (59) | Proposes an algorithm named as PrTransH to learn embedding vectors from real world EHR data based medical knowledge | Quantitative | NA | NA |
| Patienteg Dataset: Bringing Event Graph Model With Temporal Relations To Electronic Medical Records (27) | Develop a novel graph-based representation to model medical activities and temporal information from an EHR | Qualitative | TRUE | TRUE |
| Cognitive DDx Assistant In Rare Diseases (32) | Develop a cognitive solution to accelerate the differential diagnosis process by presenting information and connected knowledge to the treating physician such that medical services can be delivered faster to the patient as well as reducing the cost overhead in the health system related to rare diseases | Quantitative | NA | NA |
| Personalized Diagnostic Modal Discovery Of Traditional Chinese Medicine Knowledge Graph (94) | Develop personalized TCM KGs | Both | NA | NA |
| T-Know: A Knowledge Graph-based Question Answering And Information Retrieval System For Traditional Chinese Medicine (28) | Develop a novel knowledge service system based on the knowledge graph of TCM which includes a TCM KG, a TCM Q&A module, and a TCM knowledge retrieval module | None | NA | TRUE |

| Article | Objective[a] | Evaluation Type | Availability | |
|---|---|---|---|---|
| | | | Code | UI |
| Robokop: An Abstraction Layer And User Interface For Knowledge Graphs To Support Question Answering Bioinformatics (63) | Describe ROBOKOP and focus on capabilities enabled by the ROBOKOP user interface | None | TRUE | TRUE |
| QAnalysis: A Question-answer Driven Analytic Tool On Knowledge Graphs For Leveraging Electronic Medical Records For Clinical Research (29) | Design a novel tool, which allows doctors to enter a query using their natural language and receive their query results with charts and tables | Both | TRUE | NA |
| Fostering Natural Language Question Answering Over Knowledge Bases In Oncology EHR (33) | Present a solution to help practitioners in an oncology healthcare clinical environment with an intuitive method to access stored data | Quantitative | NA | NA |
| GAMENet: Graph Augmented Memory Networks For Recommending Medication Combination (42) | Propose GAMENet, an end-to-end deep learning model that takes both longitudinal patient EHR data and drug knowledge base on DDIs as inputs and aims to generate effective and safe recommendation of medication combination | Both | TRUE | NA |
| VisAGE: Integrating External Knowledge Into Electronic Medical Record Visualization (43) | Present VisAGE, a method that enriches patient records with a KG built from external databases | Both | NA | NA |
| KnetMaps: A BioJS Component To Visualize Biological Knowledge Networks (44) | Describe KnetMaps, an interactive BioJS component to visualise integrated knowledge networks | None | TRUE | TRUE |
| BioKEEN: A Library For Learning And Evaluating Biological Knowledge Graph Embeddings (88) | Describe BioKEEN, a python library for training models that produce knowledge embeddings | Quantitative | TRUE | NA |
| Evaluation Of Knowledge Graph Embedding Approaches For Drug-drug Interaction Prediction Using Linked Open Data (49) | Evaluate several KG embedding algorithms for predicting DDIs | Quantitative | TRUE | NA |

| Article | Objective[a] | Evaluation Type | Availability | |
|---|---|---|---|---|
| | | | Code | UI |
| Neural Networks For Link Prediction In Realistic Biomedical Graphs: A Multi-dimensional Evaluation Of Graph Embedding-based Approaches (50) | To compare the performance of neural network link prediction methods compared to baseline methods with graph embeddings as input | Quantitative | TRUE | NA |
| Edge2vec: Representation Learning Using Edge Semantics For Biomedical Knowledge Discovery (95) | Describe the edge2vec model, and its validation on biomedical domain tasks | Quantitative | TRUE | NA |
| Embedding Logical Queries On Knowledge Graphs (51) | Propose a method of performing logical queries on incomplete knowledge graphs by embedding graph nodes and representing logical operations as learned geometric operations | Quantitative | TRUE | NA |
| Medical Knowledge Embedding Based On Recursive Neural Network For Multi-disease Diagnosis (52) | Propose recursive neural knowledge network (RNKN), which combines medical knowledge based on first-order logic with recursive neural network for multi-disease diagnosis | Both | NA | NA |
| Drug-drug Interaction Prediction Based On Knowledge Graph Embeddings And Convolutional-lstm Network (53) | Introduce a framework to efficiently make predictions about conjunctive logical queries—a flexible but tractable subset of first-order logic—on incomplete KGs | Quantitative | TRUE | NA |
| Drug Target Discovery Using Knowledge Graph Embeddings (54) | Introduce a novel computational approach for predicting drug target proteins | Quantitative | NA | NA |
| Linking Physicians To Medical Research Results Via Knowledge Graph Embeddings And Twitter (55) | Apply MDE to link physicians and surgeons to the latest medical breakthroughs that are shared as the research results on Twitter | Quantitative | NA | NA |
| Opa2vec: Combining Formal And Informal Content Of Biomedical Ontologies To Improve Similarity-based Prediction (48) | Produce better concept embeddings by incorporating axiom information into the calculation | Both | TRUE | NA |
| Network Embedding In Biomedical Data Science (96) | Conduct a comprehensive review of the literature on applying network embedding to advance the biomedical domain | NA | NA | NA |

| Article | Objective[a] | Evaluation Type | Availability | |
|---|---|---|---|---|
| | | | Code | UI |
| Graph Embedding On Biomedical Networks: Methods, Applications, And Evaluations (60) | Selected 11 representative graph embedding methods and conduct a systematic comparison of three important biomedical link prediction tasks: DDA prediction, DDI prediction, PPI prediction, and two node classification tasks: medical term semantic type classification, protein function prediction | Quantitative | TRUE | NA |
| Relation Prediction Of Co-morbid Diseases Using Knowledge Graph Completion(37) | Propose a tensor factorization based approach for biological KGs | Quantitative | TRUE | NA |
| Sparklis Over Pegase Knowledge Graph: A New Tool For Pharmacovigilance (34) | Present a novel approach to enhance the way pharmacovigilance specialists perform search and exploration on their data | Both | TRUE | NA |
| OWL-NETS: Transforming Owl Representations For Improved Network Inference (38) | Propose OWL-NETS, a novel computational method that reversibly abstracts OWL-encoded biomedical knowledge into a network representation tailored for network inference | Both | TRUE | TRUE |
| Applying Knowledge-driven Mechanistic Inference To Toxicogenomics (40) | Present an MechSpy, which can be used as a hypothesis generation aid to narrow the scope of mechanistic toxicology analysis | Both | TRUE | NA |
| A Knowledge Graph-based Approach For Exploring The Us Opioid Epidemic (35) | Create the ODKG -- a network of opioid-related drugs, active ingredients, formulations, combinations, and brand names | Both | TRUE | NA |
| Investigating Plausible Reasoning Over Knowledge Graphs For Semantics-based Health Data Analytics (30) | Propose the SeDan framework that integrates plausible reasoning with expressive, fine-grained biomedical ontologies | Both | NA | TRUE |
| Interpretable Graph Convolutional Neural Networks For Inference On Noisy Knowledge Graphs (39) | Provide a new formulation for GCNNs for link prediction on graph data that addresses common challenges for biomedical KGs | Both | NA | NA |
| Structured Reviews For Data And Knowledge Driven Research (45) | Propose structured review articles as knowledge graphs focused on specific disease and research questions | Qualitative | TRUE | TRUE |

| Article | Objective[a] | Evaluation Type | Availability | |
|---|---|---|---|---|
| | | | Code | UI |
| GrEDeL A Knowledge Graph Embedding Based Method For Drug Discovery From Biomedical Literatures (57) | Propose a biomedical KG embedding-based recurrent neural network method called GrEDeL, which discovers potential drugs for diseases by mining published biomedical literature | Both | TRUE | NA |
| SemaTyp: A Knowledge Graph Based Literature Mining Method For Drug Discovery (97) | Propose a biomedical KG-based drug discovery method called SemaTyP, which discovers candidate drugs for diseases by mining published biomedical literature | Quantitative | NA | NA |
| Using Predicate Information From A Knowledge Graph To Identify Disease Trajectories (98) | Examine the potential added benefit of incorporating data provenance or metadata information into their KG and hypothesized that doing so would result in better performance on machine learning tasks | Quantitative | NA | NA |
| Using A Knowledge Graph For Hypernymy Detection Between Chinese Symptoms (99) | Propose a new method which use a knowledge graph to detect Chinese symptom hypernym relationship | Quantitative | NA | NA |
| Automatic Diagnosis With Efficient Medical Case Searching Based On Evolving Graphs (17) | Propose an automatic diagnosis method based on patient records which takes into account the temporal nature of patient records | Both | NA | NA |
| Leveraging Distributed Biomedical Knowledge Sources To Discover Novel Uses For Known Drugs (56) | Propose a novel system for drug discovery which integrates disparate biomedical knowledge through the use of a KG | Both | TRUE | NA |
| Automatic Relationship Verification In Online Medical Knowledge Base: A Large Scale Study In SemMedDB (100) | Verify medical relationships in SemMedDB | Quantitative | NA | NA |
| Gene Ontology Causal Activity Modeling (Go-Cam) Moves Beyond Go Annotations To Structured Descriptions Of Biological Functions And Systems (19) | To increase the utility of GO annotations for interpretation of genome-wide experimental data, we have developed GO-CAM, a structured framework for linking multiple GO annotations into an integrated model of a biological system. We expect that GO-CAM will enable new applications in pathway and network analysis, as well as improve standard GO annotations for traditional GO-based applications | NA | TRUE | TRUE |

| Article | Objective[a] | Evaluation Type | Availability | |
|---|---|---|---|---|
| | | | Code | UI |
| Co-occurrence Graphs For Word Sense Disambiguation In The Biomedical Domain (67) | Describe an unsupervised method for word sense disambiguation by creating a knowledge graph based on co-occurrence which is used to choose between word senses found via dictionary lookup | Quantitative | NA | NA |
| Named Entity Recognition In Traditional Chinese Medicine Clinical Cases Combining BiLSTM-CRF With Knowledge Graph (68) | Propose TCMKG-LSTM-CRF model that utilizes knowledge graph information to strengthen the learning ability and recognize rare words | Quantitative | NA | NA |
| Barack's Wife Hillary: Using Knowledge Graphs For Fact-aware Language Modeling (69) | Introduce the KGLM, a neural language model with mechanisms for selecting and copying facts from a KG that are relevant to the context | Quantitative | TRUE | NA |
| Multi-task Identification Of Entities, Relations, And Coreference For Scientific Knowledge Graph Construction (65) | Introduce a multi-task setup of identifying and classifying entities, relations, and coreference clusters in scientific articles | Quantitative | TRUE | NA |
| Biomedical Term Normalization Of EHRs With UMLS (101) | Presents a novel prototype for biomedical term normalization of EHR excerpts with the UMLs Metathesaurus | Quantitative | NA | TRUE |
| Incorporating Domain Knowledge Into Medical NLI Using Knowledge Graphs (102) | Use both word embeddings and node embeddings from a KG to infer relationships such as entailment or contradiction, between a given premise and a hypothesis | Both | NA | NA |
| A New Method For Complex Triplet Extraction Of Biomedical Texts (103) | Generate triples from biomedical text to create a KG | Quantitative | NA | NA |
| Semantic Relation Extraction Aware Of N-gram Features From Unstructured Biomedical Text (70) | Perform NER and relation extraction to generate triples to construct a KG | Quantitative | NA | NA |
| Long-tail Relation Extraction Via Knowledge Graph Embeddings And Graph Convolutional Networks (66) | To improve link prediction on heavily unbalanced datasets as a proxy to RE | Quantitative | TRUE | NA |
| MedTruth: A Semi-supervised Approach To Discovering Knowledge Condition Information From Multi-source Medical Data (62) | Propose MedTruth, a method which incorporates source quality into the reliability estimation of the knowledge is acquired | Quantitative | NA | NA |

| Article | Objective[a] | Evaluation Type | Availability | |
|---|---|---|---|---|
| | | | Code | UI |
| Towards Smart Healthcare Management Based On Knowledge Graph Technology (104) | Propose a healthy diet knowledge graph construction model that promotes the development of healthcare management | Quantitative | NA | NA |
| NormCo: Deep Disease Normalization For Biomedical Knowledge Base Construction (64) | Present NormCo, a deep coherence model which considers the semantics of an entity mention, as well as the topical coherence of the mentions within a single document | Quantitative | TRUE | NA |

[a]Whenever possible, paper objectives were copied or closely paraphrased from the original manuscripts

Abbreviations: DDA, Drug-Disease Association; DDI, Drug-Drug Interaction; EHR, Electronic Health Record; GAMENet, Graph Augmented Memory Networks; GCNN, Graph Convolutional Neural Networks; GO, Gene Ontology; KG, Knowledge graph; KGLM, Knowledge Graph Language Model; MDE, Multi Distance Knowledge Graph Embeddings; MechSpy, Mechanistic Inference Framework; NA, Not Applicable; NER, Named Entity Recognition; ODKG, Opioid Drug Knowledge Graph; OWL, Web Ontology Language; OWL-NETS, Web Ontology Language NEtwork Transformation for Statistical learning; PPI, Protein-Protein Interaction; Q&A, Question & Answer; RE, Relation Extraction; RNKN, Recursive Neural Knowledge Network; ROBOKOP, Reasoning Over Biomedical Objects linked in Knowledge Oriented Pathways; SeDan, SEmantics-based Data ANalytics; SemMedDB, The Semantic MEDLINE Database; TCM, Traditional Chinese Medicine; TCMKG-LSTM-CRF, Traditional Chinese Medicine Long Short-Term Memory Conditional Random Field; UMLS, Unified Medical Language System; VisAGE, Present Visualization Assisted by Knowledge Graph Enrichment.

Table 2    Reviewed Articles on Constructing Knowledge Graphs

| Article | Objective[a] | Evaluation Type | Availability | |
|---|---|---|---|---|
| | | | Code | UI |
| Liberating Links Between Datasets Using Lightweight Data Publishing: An Example Using Plant Names And The Taxonomic Literature (71) | Describe the creation of a datasette for a longstanding but mostly unpublished project on linking plant names in The Index of Plant Names Index to the taxonomic literature | Qualitative | TRUE | TRUE |
| Ozymandias: A Biodiversity Knowledge Graph (72) | Explore the feasibility of constructing a "biodiversity knowledge graph" for the Australian fauna | Qualitative | TRUE | TRUE |
| OpenBiodiv-O: Ontology Of The OpenBiodiv Knowledge Management System (73) | Introduce OpenBiodiv-O, an ontology that fills the gaps between ontologies for biodiversity resources, such as DarwinCore-based ontologies, and semantic publishing ontologies, such as the SPAR Ontologies | Qualitative | TRUE | TRUE |
| Challenges In The Construction Of Knowledge Bases For Human Microbiome-disease Associations (74) | Survey the existing tools and development efforts that have been produced to capture portions of the information needed to construct a KG of all known human microbiome-disease associations and highlight the need for additional innovations in NLP, text mining, taxonomic representations, and field-wide vocabulary standardization in human microbiome research | None | NA | NA |
| Construction Of The Literature Graph In Semantic Scholar (105) | Describe a deployed scalable system for organizing published scientific literature into a heterogeneous graph to facilitate algorithmic manipulation and discovery | Quantitative | TRUE | NA |
| Towards A Knowledge Graph For Science (106) | Propose the vision of a knowledge graph for science and present a possible infrastructure for building such a KG | None | NA | NA |
| Distantly Supervised Biomedical Knowledge Acquisition Via Knowledge Graph Based Attention (107) | Proposed a new end-to-end KGC model | Quantitative | NA | NA |
| The Role Of: A Novel Scientific Knowledge Graph Representation And Construction Model (108) | Propose a novel representation of SciKG, which has three layers. The first layer has concept nodes, attribute nodes, as well as the attaching links from attribute to concept. The second layer represents both fact | Both | NA | NA |

| Article | Objective[a] | Evaluation Type | Availability | |
|---|---|---|---|---|
| | | | Code | UI |
| | tuples and condition tuples | | | |
| Mining Scholarly Publications For Scientific Knowledge Graph Construction (109) | Present a preliminary approach that uses a set of NLP techniques for extracting entities and relationships from research publications and then integrated them in a KG | Quantitative | NA | NA |
| Biological Knowledge Graph Construction, Search, And Navigation (110) | Illustrate how a knowledge graph can be constructed out of the biological RDF databases | None | NA | NA |
| Constructing Large Scale Biomedical Knowledge Bases From Scratch With Rapid Annotation Of Interpretable Patterns (81) | Present a simple and effective method of extracting new facts with a pre-specified binary relationship type from the biomedical literature, without requiring any training data or hand-crafted rules | Quantitative | NA | NA |
| Constructing Biomedical Domain-specific Knowledge Graph With Minimum Supervision (80) | Propose a versatile approach for knowledge graph construction with minimum supervision based on unstructured biomedical domain-specific contexts including the steps of entity recognition, unsupervised entity and relation embedding, latent relation generation via clustering, relation refinement and relation assignment to assign cluster-level labels | Both | NA | NA |
| Constructing Biomedical Knowledge Graph Based On SemMedDB And Linked Open Data (82) | Propose a novel approach to building a BMKG leveraging the SemMedDB and Health Science LOD | Both | NA | NA |
| Towards FAIRer Biological Knowledge Networks Using A Hybrid Linked Data And Graph Database Approach (83) | Present our developments to connect, search and share data about genome-scale knowledge networks | Quantitative | NA | NA |
| Construction Of Traditional Chinese Medicine Knowledge Graph Using Data Mining And Expert Knowledge (76) | Propose to construct a knowledge graph of individual TCM expertise through data mining on a small dataset of EHRs, which leverages both the availability of some small datasets and the analysis from experts by saving their conclusions into KG | Both | NA | NA |

| Article | Objective[a] | Evaluation Type | Availability | |
|---|---|---|---|---|
| | | | Code | UI |
| Integrating Biomedical Research And Electronic Health Records To Create Knowledge Based Biologically Meaningful Machine Readable Embeddings (75) | Describe a method for embedding clinical features from EHRs onto SPOKE | Both | TRUE | TRUE |
| On Building A Diabetes Centric Knowledge Base Via Mining The Web (111) | Propose an approach to constructing a DKB via mining the Web | Both | NA | NA |
| Personalized Health Knowledge Graph (112) | Explain the research challenges to designing the PHKG and provide our vision to address known challenges to constructing these types of KGs | None | TRUE | NA |
| Design And Implementation Of Personal Health Record Systems Based On Knowledge Graph (113) | We propose a novel approach to design the knowledge graph of medical information and to implement an effective PHR system to support individual healthcare management | Qualitative | NA | NA |
| Intelligent Healthcare Knowledge Resources In Chinese: A Survey (114) | Surveys the medical resources needed to construct a medical knowledge base and ways to obtain medical resources | None | NA | NA |
| Modelling Online User Behavior For Medical Knowledge Learning (115) | Propose a TLLFG, which can learn the descriptions of ailments used in internet searches and match them to the most appropriate formal medical keywords | Both | NA | NA |
| Building Causal Graphs From Medical Literature And Electronic Medical Records (116) | Novel approach for automatically constructing causal graphs between medical conditions | Both | TRUE | NA |
| Mining Disease-symptom Relation From Massive Biomedical Literature And Its Application In Severe Disease Diagnosis (117) | Present a study on mining disease-symptom relation from massive biomedical literature and constructing biomedical KG from the relation | Both | NA | NA |
| OC-2-KB: Integrating Crowdsourcing Into An Obesity And Cancer Knowledge Base Curation System (78) | An update to OC-2-KB. It is a software pipeline which automatically extracts semantic triples from PubMed abstracts related to obesity and cancer | Both | TRUE | NA |
| Diagnosis Of COPD Based On A Knowledge Graph And Integrated Model (79) | Construct a KG of COPD to mine and better understand the relationship between diseases, symptoms, causes, risk factors, drugs, side effects, and more | Both | NA | NA |

| Article | Objective[a] | Evaluation Type | Availability | |
|---|---|---|---|---|
| | | | Code | UI |
| Aero: An Evidence-based Semantic Web Knowledge Base Of Cancer Behavioral Risk Factors (77) | Present a prototype, Aero to (1) better organize and provide evidence based CBRF knowledge extracted from scientific literature (i.e., PubMed), and (2) provide users with access to high-quality scientific knowledge, yet easy to understand answers for their frequently encountered CBRF questions | Both | NA | NA |
| ALOHA: Developing An Interactive Graph-based Visualization For Dietary Supplement Knowledge Graph Through User-centered Design (118) | Improved a novel interactive visualization platform, ALOHA, for the general public to obtain DS-related information through two user-centered design iterations | Both | TRUE | NA |
| Separating Wheat From Chaff: Joining Biomedical Knowledge And Patient Data For Repurposing Medications (119) | Present a system that jointly harnesses large-scale electronic health records data and a concept graph mined from the medical literature to guide drug repurposing—the process of applying known drugs in new ways to treat diseases | Both | NA | NA |
| KGDDS: A System For Drug-drug Similarity Measure In Therapeutic Substitution Based On Knowledge Graph Curation (120) | Present a system KGDDS for node-link-based bio-medical KG curation and visualization, aiding Drug-Drug Similarity measure | Both | TRUE | TRUE |
| Supporting Shared Hypothesis Testing In The Biomedical Domain (121) | Propose a method of assembling biological knowledge about biological processes into hypothesis graphs, a computational framework for hypothesis testing | Both | TRUE | TRUE |

[a]Whenever possible, paper objectives were copied or closely paraphrased from the original manuscripts

Abbreviations: Aero, cAncer bEhavioral Risk knOwledge base; BMKG, Biomedical Knowledge Graph; CBRF, Cancer Behavioral Risk Factors; COPD, Chronic Obstructive Pulmonary Disease; DKB, Diabetes Centric Knowledge Base; EHR, Electronic Health Record; KG, Knowledge Graph; KGC, Knowledge Graph Representation And Construction Model; KGDDS, KG Drug-Drug Similarity; LOD, Linked Open Data; NA, Not Applicable; NLP, Natural Language Processing; OC-2-KB, Obesity and Cancer to Knowledge Base; PHKG, Personalized Health Knowledge Graph; PHR, Personal Health Record; RDF, Resource Description Framework; SciKG, Scientific Knowledge Graph Representation And Construction Model; SemMedDB, Semantic MEDLINE Database; SPAR, Semantic Publishing and Referencing Ontologies; SPOKE, Scalable Precision Medicine Oriented Knowledge Engine; TCM, Traditional Chinese Medicine; TLLFG, Transfer Learning using Latent Factor Graph.

Table 3 Currently available biomedical data science knowledge graphs

| Name | Primitives | Domain | Endpoint | Last Updated | Construction Method |
|---|---|---|---|---|---|
| Bio2RDF (122) | Ontology concepts (URIs) | Biomedical | Virtuoso | 09/25/14 | Semantic integration of OBOs |
| BioGrakn (123) | Ontology concepts (URIs) | Biomedical | Grakn.ai | 09/27/19 | Semantic integration of OBOs |
| DisGeNET (124) | Ontology concepts (URIs) | Biomedical | Unspecified | 01/01/19 | Semantic integration of OBOs |
| HetioNet (125) | Ontology concepts (URIs) | Biomedical | Neo4J | 07/08/19 | Semantic integration of OBOs |
| KaBOB (46) | Ontology concepts (URIs) | Biomedical | Allegrograph | 06/23/19 | Semantic integration of OBOs |
| NGLY1 Deficiency (45) | Ontology concepts (URIs) | NGLY1 Deficiency | Neo4J | 08/08/19 | Semantic integration of OBOs |
| Ozymandias (72) | Ontology concepts (URIs) | Biodiversity | Blazegraph | 2019 | Linked data from ALA and ALD using CrossRef |
| PheKnowLator (61) | Ontology concepts (URIs) | Human Biomedical | None | 09/25/19 | Semantic integration of OBOs |
| ROBOKOP (63) | Ontology concepts (URIs) | Biomedical | GreenT Neo4J | 09/20/19 | Semantic integration of OBOs |
| Sparklis (34) | Ontology concepts (URIs) | Pharmacovigilance | SPARQL | 01/2019 | Semantic integration of OBOs |

Abbreviations: ALD, Australian Faunal Directory; ALA, Atlas of Living Australia; NGLY1, N-glycanase 1; OBO, Open Biomedical Ontology; URI, Universal Resource Identifier.

# SUPPLEMENTAL MATERIAL

Supplemental Table 1 Reviewed Articles on Applications of KGs in Biomedical Data Science

| Article | Topic | Objective[a] | Approach | | | Evaluation | | | Contributions[a] | Publicly Available | |
|---|---|---|---|---|---|---|---|---|---|---|---|
| | | | Input Data | Overview | Tools | Type | Strategy | Metric(s) | | Code | Interface |
| Exploiting Semantic Patterns Over Biomedical Knowledge Graphs For Predicting Treatment And Causative Relations (31) | Disease Diagnosis | Build high accuracy supervised predictive models to predict previously unknown treatment and causative relations between biomedical entities based only on semantic graph pattern features extracted from biomedical KGs | UMLS, SemMedDB | Built a KG using SemMedDB and manually curated a specific set of 7,000 "treats" and 2,918 "causes" relations from the UMLS | Logistic regression, Decision trees | Both | Applied supervised machine learning models to predict "treats" and "causes" relations from graph patterns connecting biomedical entities<br><br>Validation: Applied model to repoDB. Two clinical domain experts reviewed results | Mean precision, Mean recall, Mean F1-score | Predicted treats/causes relations with high F-scores of 90-99% and 90%<br><br>Retrieved >50% of treats relations when applied to an external dataset<br><br>Identified new plausible relations verified by two physician co-authors | TRUE | TRUE |
| Retrieval Method Of Electronic Medical Records Based On Rules And Knowledge Graph (26) | General | Demonstrate how KGs improve the retrieval of information from electronic health records | Shengjing Hospital EHR | Developed an information retrieval system that leverages a KG built from an EHR. The system consists of 4 steps: reconstruct query using KG, perform first search and generate results, perform second search on results using rules and dependency parsing, and return final result | RE, Attribute extraction, Coreference resolution, Entity disambiguation | Quantitative | Performed eight different information retrieval queries on a subset of 30,000 records from Shengjing Hospital EHR related to lung disease | Mean average precision, Binary preference, Top-100 Precision | Combining a rules-based approach with a KG resulted in better performance than using only a rules-based approach | NA | NA |
| PrTransH: Embedding Probabilistic Medical Knowledge From Real World EHR Data (59) | General | Proposes an algorithm named as PrTransH to learn embedding vectors from real world EHR data based medical knowledge | Anhui Medical University EHR data, ICD 10 | KG constructed from EHR, nodes were ICD 10 diagnoses and edges types were: disease_to_medicine, disease_to_symptom, disease_to_operation, disease_to_laboratory, disease_to_examination, disease_to_lower_disease. Developed a novel probabilistic algorithm that leverages KG | Link prediction | Quantitative | Performed link prediction on KG constructed from 3,767,198 patients and 16,217,270 visits. Compared performance of algorithm to TransH | Mean rank, Hits@10, NDCG@10 | PrTransH is able to successfully incorporate and embed uncertainty of diagnoses into EHR-based embeddings<br><br>PrTransH performs significantly better than TransH on link prediction task for all metrics | NA | NA |

| Article | Topic | Objective[a] | Approach | | | Evaluation | | | Contributions[a] | Publicly Available | |
|---|---|---|---|---|---|---|---|---|---|---|---|
| | | | Input Data | Overview | Tools | Type | Strategy | Metric(s) | | Code | Interface |
| Patienteg Dataset: Bringing Event Graph Model With Temporal Relations To Electronic Medical Records (27) | Personal KG | Develop a novel graph-based representation to model medical activities and temporal information from an EHR | Shanghai Shuguang Hospital EHR | Each patient event graph was constructed in 4 steps: data preprocessing, generate event triples generation, establish temporal relations, and match clinical concepts from EHR to entities in CBioMedKG | SEMI, RDB2RDF, Levenshtein Similarity, Jaccard Similarity, Longest Common Subsequence Similarity | Qualitative | Built a SPARQL interface and provided queries to help users explore KG representations | NA | Constructed a PatientEG dataset from EHRs based on the proposed model, and were able to link entities in CBioMedKG enabling users to construct queries on patients using domain knowledge<br><br>Publish linked PatientEG dataset and provide online access (with queries) via SPARQL endpoint | TRUE | TRUE |
| Cognitive DDx Assistant In Rare Diseases (32) | Rare Disease | Develop a cognitive solution to accelerate the differential diagnosis process by presenting information and connected knowledge to the treating physician such that medical services can be delivered faster to the patient as well as reducing the cost overhead in the health system related to rare diseases | ICD-10, medRA, DOID, PubMed Abstracts, CDC.gov, MeSH, Wikipedia, Orphanet, Anonymized patient data | Built a proprietary KG from input sources | IBM Research tools | Quantitative | Measured success retrieving correct patients using 101 queries defined by prior work. Determined if removing specific data sources results in improved performance | Hits @1%, Overall average of Hits@1% | 61% hit rate for initial build using all data sources, which improve to 79.5% after two additional sprints and removing a subset of Orphanet data<br><br>Using Wikipedia alone achieved a 74% success rate, but some of the rare diseases were excluded | NA | NA |
| Personalized Diagnostic Modal Discovery Of Traditional Chinese Medicine Knowledge Graph (94) | TCM | Develop personalized TCM KGs | TCM KG, TCM EHR data | First, EHR medical records were standardized by applying a measure of semantic similarity to an existing TCM KG. Then, path queries for medical cases are run by identifying all symptoms-syndrome, syndrome-treatment, and treatment-chinese medicine connections. Finally, a personalized KG is built by superimposing and traversing the matrices that are built from the stored path queries | Matrix search | Both | 102 sets of data from a TCM doctor were evaluated by examining the main symptoms, identifying of intermediate symptom and syndrome nodes, and verifying diagnosis mode by running reasoner.<br><br>Validation: TCM heritage team verification of results | None | Successfully developed and tested a framework for creating a TCM personalized KG | NA | NA |

| Article | Topic | Objective[a] | Approach | | | Evaluation | | | Contributions[a] | Publicly Available | |
|---|---|---|---|---|---|---|---|---|---|---|---|
| | | | Input Data | Overview | Tools | Type | Strategy | Metric(s) | | Code | Interface |
| T-Know: A Knowledge Graph-based Question Answering And Information Retrieval System For Traditional Chinese Medicine (28) | TCM | Develop a novel knowledge service system based on the KG of TCM which includes a TCM KG, a TCM Q&A module, and a TCM knowledge retrieval module | EHR data, Clinical guidelines, Teaching materials, Medical books, Publications | KG: Build a TCM KG from the described sources using the input data and named entity recognition and RE. The KG contains diseases, symptoms, syndromes, prescriptions, and chinese herbals.<br><br>Q&A: Input sources were processed using medical named entity reorganization and relation extraction, entity linking, joint disambiguation of entities and relations<br><br>Knowledge Retrieval: consisted of search word extension and human-computer interactive retrieval | BiLSTM+CRF, CNN, Maximum Sampling, S-MART | None | NA | NA | Proposed and described a novel TCM information retrieval system called T-Know<br><br>Built a TCM KG that contained >10,000 nodes and 220,000 edges | NA | TRUE |
| Robokop: An Abstraction Layer And User Interface For Knowledge Graphs To Support Question Answering Bioinformatics (63) | Medical Q&A | Describe ROBOKOP and focus on capabilities enabled by the ROBOKOP user interface | ROBOKOP KG, OmniCorp maintained PubMed identifiers and linked concepts | Designed a system to return query results in the form of sub-graphs. When multiple results exist, the most relevant sub-graph is determined by weighting each edges according to a confidence score which is derived using counts from curated data sources as well as co-occurrence counts from PubMed abstracts | Resistance distance | None | NA | NA | A novel technology stack which includes a biomedical KG, API, and user interface | TRUE | TRUE |

| Article | Topic | Objective[a] | Approach | | | Evaluation | | | Contributions[a] | Publicly Available | |
|---|---|---|---|---|---|---|---|---|---|---|---|
| | | | Input Data | Overview | Tools | Type | Strategy | Metric(s) | | Code | Interface |
| QAnalysis: A Question-answer Driven Analytic Tool On Knowledge Graphs For Leveraging Electronic Medical Records For Clinical Research (29) | Clinical Enrichment | Design a novel tool, which allows doctors to enter a query using their natural language and receive their query results with charts and tables | EHR data, Chinese clinical terminology graph, Chinese medical dictionaries | KG: Built to represent symptoms, diseases, and tests from a clinical CDM. Patient data is represented using a data, schema, and terminology KG<br><br>Q&A: word segmentation, concept linking, grammar parsing, semantic representation and concept disambiguation, and finally Cypher query language translation were applied to process questions | OMOP CDM, Stanford parser, Jieba word segmenter, Levenshtein distance | Both | Gold-standard questions: collected from domain experts and the literature. Domain expert questions (n=50) were collected by performing clinical interviews in addition to clinician-ranked questions via website. Questions were converted to a tree-like knowledge representation. Questions from the literature were obtained by querying a chinese search engine and the CALIBER research project (n=161 questions from 28 papers)<br><br>Data: Congestive heart failure patients was used to build a KG of 421,728 nodes and 86,525 edges<br><br>Validation: Gold-standard questions were reviewed by a team of clinicians and IT professionals | Coverage, Precision, Query time | Design a graph-based patient-specific schema, which can be extended or revised according to fit different application contexts<br><br>Collected sets of analytic questions from doctors and designed a tree-like knowledge representation to represent them, which included different kinds of operators used in clinical research | TRUE | NA |

| Article | Topic | Objective[a] | Approach | | | Evaluation | | | Contributions[a] | Publicly Available | |
|---|---|---|---|---|---|---|---|---|---|---|---|
| | | | Input Data | Overview | Tools | Type | Strategy | Metric(s) | | Code | Interface |
| Fostering Natural Language Question Answering Over Knowledge Bases In Oncology EHR (33) | Oncology | Present a solution to help practitioners in an oncology healthcare clinical environment with an intuitive method to access stored data | InterProcess Gemed Oncology EHR data | Data: 3,309 unstructured EHR notes were anonymized and exported as XML documents. Focus group-generated sentences were used to assist in extracting meaningful information from the XML documents<br><br>KG: Focus-group sentences were used to develop KG schema<br><br>Q&A: Identified Spotlight and the Google Knowledge Graph in order to identify question types and keywords. Named entity recognition was performed using the Palavras Parser | ENSEPRO | Quantitative | Evaluation is performed via scenario-based assessment. | Precision | Developed an oncology ontology composed of 53 classes and 12 relations<br><br>Outlined the application of the developed system for question answering over the KG | NA | NA |

| Article | Topic | Objective[a] | Approach | | | Evaluation | | | Contributions[a] | Publicly Available | |
|---|---|---|---|---|---|---|---|---|---|---|---|
| | | | Input Data | Overview | Tools | Type | Strategy | Metric(s) | | Code | Interface |
| GAMENet: Graph Augmented Memory Networks For Recommending Medication Combination (42) | Drug-drug Interactions | Propose GAMENet, an end-to-end deep learning model that takes both longitudinal patient EHR data and drug knowledge base on DDIs as inputs and aims to generate effective and safe recommendation of medication combination | MIMIC-III, TWOSIDES | Patient representation: Use separate RNNs to separately encoded diagnoses and procedures for each visit<br><br>Medical embeddings: Use binary encoded multi-hot vectors to encode each medical concept that occurs at a given visit<br><br>Drug-Drug KG: built by refining input data to only include the top-40 most severe drug-drug interactions<br><br>FInally, a graph augmented memory module is created from the emnbedded patient EHR data and the drug-drug interaction graph by treating the KG and embedding data as facts in a memory bank as well as using the patient history to establish Dynamic Memory | RNN, DM, MB, GCNs, Multi-label loss | Both | Data: Patients with more than 1 visit and only use medication data from the first 24-hours<br><br>Baselines: Nearest, logistic regression, LEAP, RETAIN, DMNC<br><br>Examined a single complicated case study in order to provide an explanation of the poor performance of the baseline methods | DDI Rate, Jaccard Similarity, Precision Recall AUC, Average F1-Score | Presented GAMENet, an end-to-end deep learning model that aims to generate effective and safe recommendations of medication combinations via memory networks whose memory bank is augmented by integrated drug usage and DDI graphs as well as dynamic memory based on patient history<br><br>Experimental results showed that GAMENet outperformed all baselines in effectiveness measures, and achieved a 3.60% DDI rate reduction | TRUE | NA |
| VisAGE: Integrating External Knowledge Into Electronic Medical Record Visualization (43) | Parkinson's Disease | Present VisAGE, a method that enriches patient records with a KG built from external databases | Parkinson's Progression Markers Initiative data, inBioMap, STITCH | Data: Derived a patient profile matrix, where rows correspond to patients and columns correspond to features. The matrix was built using 1,579 patients and 6,013 features<br><br>KG: Constructed from input data sources, including SNPs, EHR-derived symptom-drug co-occurrence. Embeddings were generated and similar concepts were then connected to concepts in the patient profile matrix | ProSNet, Cosine similarity | Both | Use t-SNE to visualize the patient profile matrix with and without enrichment from KG embeddings. Clustering was then performed and the enrichment of specific drugs and symptoms within the resulting clusters was measured using significance tests | Unified Parkinson's disease rating scale score, Fisher's exact test, BH procedure | Created a KG which contained 23,886 nodes and 17,108,116 edges<br><br>Qualitatively and quantitatively show that VisAGE can more effectively cluster patients, which allows doctors to better discover patient subtypes and thus improve patient care | NA | NA |

| Article | Topic | Objective[a] | Approach | | | Evaluation | | | Contributions[a] | Publicly Available | |
|---|---|---|---|---|---|---|---|---|---|---|---|
| | | | Input Data | Overview | Tools | Type | Strategy | Metric(s) | | Code | Interface |
| KnetMaps: A BioJS Component To Visualize Biological Knowledge Networks (44) | Biological Relationships | Describe KnetMaps, an interactive BioJS component to visualise integrated knowledge networks | KnetMiner | KnetMaps was built on top of Cytoscape and jQuery and requires data be input in JSON. The application was customized to facilitate a wide variety of interaction capabilities. Nodes and edges are visualized together using a force-directed layout and labels for both nodes and edges can be customized. A custom Item Information panel, which allows the user the ability to explore node and edge relations as key-value pairs was developed. The application is able to accommodate networks with up to 1,000 nodes and 3,000 edges and is currently used by KnetMiner and Daisychain | Force-directed layout | None | NA | NA | KnetMaps is a fast and lightweight touch-friendly tool for visualizing content-rich, heterogeneous knowledge networks<br><br>The implementation uses cytoscapeJS, jQuery and JavaScript extensions for interactive functionality to ensure that low-memory, touch-compatible networks can be rendered in web browsers without the need to write extensive and unwieldy server-side code | TRUE | TRUE |
| BioKEEN: A Library For Learning And Evaluating Biological Knowledge Graph Embeddings (88) | KG Embedding Evaluation | Describe BioKEEN, a python library for training models that produce knowledge embeddings | ADEPTUS, ComPath, DrugBank, ExPASy, HIPPIE, HSDN, KEGG, mirTarBase, MSigDB, Reactome, InterPro, WikiPathways | BioKEEN is a python library built on top of PyKEEN. It provides ten models and tools for configuring, exploring, and optimizing model parameters. | Bio2BEL, TransE, TransH, TransR, TransD, ConvE, SE, UM, RESCAL, DistMult, ERMLP | Quantitative | Generate triples from Bio2BEL and user-provided sources<br><br>Perform hyperparameter optimization from a random search grid, and generate negative samples before training the model to generate embeddings | Mean rank<br><br>Hits@k | Provided a library for training and tuning models for producing knowledge embeddings. This is a methods paper, no novel biomedical finding | TRUE | NA |
| Evaluation Of Knowledge Graph Embedding Approaches For Drug-drug Interaction Prediction Using Linked Open Data (49) | Drug-drug Interactions | Evaluate several KG embedding algorithms for predicting DDIs | Bio2RDF | Using a KG built from entities and relations derived from DrugBank, a variety of tools were used to produce knowledge embeddings, then predicted DDIs using three classification methods | RDF2Vec, TransE, TransD | Quantitative | The knowledge embedding methods were evaluated by evaluating how well classification methods performed when the embeddings were used as input | AUC, F1-Score, AUPR | RDF2Vec with uniform weighting outperformed other methods | TRUE | NA |

| Article | Topic | Objective[a] | Approach | | | Evaluation | | | Contributions[a] | Publicly Available | |
|---|---|---|---|---|---|---|---|---|---|---|---|
| | | | Input Data | Overview | Tools | Type | Strategy | Metric(s) | | Code | Interface |
| Neural Networks For Link Prediction In Realistic Biomedical Graphs: A Multi-dimensional Evaluation Of Graph Embedding-based Approaches (50) | DTI, PPI | To compare the performance of neural network link prediction methods compared to baseline methods with graph embeddings as input | MATADOR, BioGRID, PubTator | Computed graph embeddings from the input graphs using LINE, DeepWalk, SDNE, and Node2Vec. Then used a neural link predictor as well as three baseline methods to make link predictions | DeepWalk, LINE, Node2vec, SDNE, Adamic-Adar, Common Neighbors, Jaccard Index | Quantitative | Randomly sliced input graphs, using 60% of the links to create graph embeddings, 10% to train the link predictor, and 40% set aside to evaluate the performance. | AUROC, AUPRC, Average R-precision, Precision@k | Approaches were equal among all method at low recall levels, but the neural network approaches were superior at high recall levels | TRUE | NA |
| Edge2vec: Representation Learning Using Edge Semantics For Biomedical Knowledge Discovery (95) | Knowledge Discovery | Describe the edge2vec model, and its validation on biomedical domain tasks | Chem2Bio2RDF | Use an edge-type transition matrix to represent network heterogeneity. Use an EM model to train a transition matrix via random walks. Embeddings are then learned using stochastic gradient descent | EM, edge2VEC, Random walk, SGD | Quantitative | The edge2vec model was trained using Chem2Bio2RDF and performed three tasks: entity multi-classification, compound-gene bioactivity prediction, and compound-gene search ranking. It was compared to 3 other methods: DeepWalk, LINE, and node2vec. Parameter tuning of edge2vec was also explored | Entity multi-classification, Compound-gene bioactivity prediction, Compound-gene search ranking | edge2vec significantly outperformed existing knowledge embedding models on three biomedical domain tasks<br><br>IncorporatIng edge-types in the node embedding learning process for heterogeneous graphs is important | TRUE | NA |
| Embedding Logical Queries On Knowledge Graphs (51) | DGDI | Propose a method of performing logical queries on incomplete KGs by embedding graph nodes and representing logical operations as learned geometric operations | STITCH, SIDER, OFFSIDES, DisGeNET, Reddit | Generate both node and query embeddings. Query embeddings are generated by first calculating a geometric projection operator and a geometric intersection operator and algorithmically computing an embedding that corresponds to a conjunctive query | Bilinear Bag-of-features, TransE, DistMult | Quantitative | For a test query, compare how well the model ranks a node that does satisfy the query to nodes that do not satisfy the query. Bot a Bio data graph and a Reddit derived graph were used | AUROC | Proposed a framework to embed conjunctive graph queries, demonstrating how to map a practical subset of logic to efficient geometric operations in an embedding space<br><br>Can make accurate predictions on real-world data with millions of relations | TRUE | NA |

| Article | Topic | Objective[a] | Approach | | | Evaluation | | | Contributions[a] | Publicly Available | |
|---|---|---|---|---|---|---|---|---|---|---|---|
| | | | Input Data | Overview | Tools | Type | Strategy | Metric(s) | | Code | Interface |
| Medical Knowledge Embedding Based On Recursive Neural Network For Multi-disease Diagnosis (52) | Disease Diagnosis | Propose recursive neural knowledge network (RNKN), which combines medical knowledge based on first-order logic with recursive neural network for multi-disease diagnosis | CEMRs | Used an architecture based on RNNs.<br><br>Symptom and disease embeddings are the input layer and medical knowledge base determines connection between the input layer and the shallow logic layer<br><br>A Huffman tree based on the frequency of knowledge is used as the deep layers<br><br>The model is then trained using gradient descent and back-propagation | RNKN, RNN, Softmax | Both | Evaluated effects of epochs and dimensionality of the embeddings<br><br>Compared to machine learning models and Markov logic networks in multi-disease diagnosis task<br><br>Assessed interpretability of embeddings using t-SNE plots to visually find communities and labelling with ICD 10 codes | t-SNE, Community identification, Precision @k, DCG | RNKN outperformed other methods when used for disease diagnosis tasks<br><br>Knowledge embeddings can be interpreted using t-SNE plots and community detection<br><br>Mathematical derivations of back-propagation and forward-propagation were deduced | NA | NA |

| Article | Topic | Objective[a] | Approach | | | Evaluation | | | Contributions[a] | Publicly Available | |
|---|---|---|---|---|---|---|---|---|---|---|---|
| | | | Input Data | Overview | Tools | Type | Strategy | Metric(s) | | Code | Interface |
| Drug-drug Interaction Prediction Based On Knowledge Graph Embeddings And Convolutional-lstm Network (53) | Drug-drug Interaction | Introduce a framework to efficiently make predictions about conjunctive logical queries—a flexible but tractable subset of first-order logic—on incomplete KGs | DrugBank, KEGG, OFFSIDES, PharmGKB | DDI prediction from a KG that combined just a few ontologies using a variety of concept embeddings. They devised their own embedding strategy using a combination of CNNs and LSTMs. The DDI information was curated from DrugBank, KEGG, TWOSIDES and MEDLINE (rather than a single source like NLM's RXNorm) | Convolutional-LSTM, (CNN + LSTM) | Quantitative | Compiled a list of DDIs from public databases.<br><br>Created KG from mentioned sources and calculated node embeddings.<br><br>Feed the concatenated embeddings of each drug to a convolution layer, then to an LSTM to generate the prediction. Other features derived from DrugBank, etc were used to compare against other M models as a baseline (SVM, KNN, etc) | AUPRC, Positive fraction / Mean predicted value | Created a dataset with 2,898,937 drug-drug interaction pairs; believe that this is the largest available<br><br>Prepared a large-scale integrated KG about DDIs with data from DrugBank, KEGG, OFFSIDES, and PharmGKB having 1.2 billion triples<br><br>Evaluated different KG embeddings techniques with different settings to train and evaluate ML models<br><br>Provide a comprehensive evaluation with details analysis of the outcome and comparison with the state-of-the-art approaches and baseline models<br><br>Found that a combined CNN and LSTM network called Conv-LSTM for predicting DDIs leads to the highest accuracy | TRUE | NA |

| Article | Topic | Objective[a] | Approach | | | Evaluation | | | Contributions[a] | Publicly Available | |
|---|---|---|---|---|---|---|---|---|---|---|---|
| | | | Input Data | Overview | Tools | Type | Strategy | Metric(s) | | Code | Interface |
| Drug Target Discovery Using Knowledge Graph Embeddings (54) | Drug Target Discovery | Introduce a novel computational approach for predicting drug target proteins | KEGG | Used a subset of KEGG as their KG to predict gene targets for drugs, using a custom model to generate node and edge embeddings. From these, they attempt to predict new edges representing novel targets | ComplEx | Quantitative | Generated embeddings by both training on correct graph triples and by "corrupting" the KG adding fake triples (to generate negative samples)<br><br>Predict links for any given drug, as well as for any given gene, and those that match on each end are considered as predicted drug-gene relations | MRR, Hits@k | Introduced the use of KG embedding models for predicting drug targets using currently available drug knowledge bases<br><br>Created a KG dataset, KEGG50k, from KEGG database, which is centred around drugs and their targeted genes, disease and reaction pathways. Used this dataset to evaluate the predictive accuracy of KG embedding models | NA | NA |
| Linking Physicians To Medical Research Results Via Knowledge Graph Embeddings And Twitter (55) | Literature Suggestion System | Apply MDE to link physicians and surgeons to the latest medical breakthroughs that are shared as the research results on Twitter | Twitter data | They generated a KG from a large dataset of Twitter data, identifying users as researchers, physicians, etc. This is another link prediction task | MDE | Quantitative | Create KG from Twitter data retrieved, using user profile metadata to infer user categories<br><br>Create edge embeddings to be able to predict several relation types (is_talking_about, job_title_type_is, etc) | MRR, Hits@k | Proposed the usage of MDE to suggest Tweets about medical breakthroughs to physicians<br><br>Experiment shows the superior ranking performance of MDE over the baseline<br><br>Model can be suggested to serve in connecting the physicians and the up-to-date advances in the medical studies | NA | NA |
| Opa2vec: Combining Formal And Informal Content Of Biomedical Ontologies To Improve Similarity-based Prediction (48) | Concept Enhancement | Produce better concept embeddings by incorporating axiom information into the calculation | GO/GOA, PhenomeNET, STRING, HPO, MGI, OMIM, DOID, PubMed full text, MEDLINE abstracts | This study combined concept embeddings with word embeddings from some annotations (label, description, synonyms, etc) for said concept, to produce a more informative vector representation. This was validated in a PPI task, using STRING as ground truth. The model trained on fulltext performs better than that trained on abstracts | ELK, OWL API | Both | Pre-train Word2vec model on MEDLINE abstracts and PubMed full text<br><br>Create a KG from GO and protein instances (one KG for humans, another for yeast proteins)<br><br>Train a deep neural net to combine these embeddings<br><br>The PPI and gene-disease associations are determined based on semantic similarity | Cosine similarity, Resnik semantic similarity, ROC, AUC | Demonstrated that OPA2Vec can significantly improve predictive performance in applications that rely on the computation of semantic similarity<br><br>Results illustrate that the annotation properties that are used to describe details about an ontology class in natural language, in particular the labels and descriptions, contribute most to the feature vectors<br><br>Exploits use of ontologies as community standards, and inclusion of both human- and machine-readable information in ontologies as standard requirements for publishing ontologies | TRUE | NA |

| Article | Topic | Objective[a] | Approach | | | Evaluation | | | Contributions[a] | Publicly Available | |
|---|---|---|---|---|---|---|---|---|---|---|---|
| | | | Input Data | Overview | Tools | Type | Strategy | Metric(s) | | Code | Interface |
| Network Embedding In Biomedical Data Science (96) | KG Embedding Review | Conduct a comprehensive review of the literature on applying network embedding to advance the biomedical domain | NA | An overview of network embedding methods is provided as well as descriptions of applications of these methods in biomedical data science. Finally a detailed overview of the challenges to utilizing these methods is provided | NA | NA | NA | NA | Results from reviewed literature illustrate the capabilities of network embeddings for biomedical network analysis<br><br>Summarized the challenges of using network embedding applications within the biomedical domain | NA | NA |
| Graph Embedding On Biomedical Networks: Methods, Applications, And Evaluations (60) | KG Embedding Evaluation | Selected 11 representative graph embedding methods and conduct a systematic comparison of three important biomedical link prediction tasks: DDA prediction, DDI prediction, PPI prediction, and two node classification tasks: medical term semantic type classification, protein function prediction | CTD, NDF-RT, DrugBank, STRING, EHR data, UMLS | Provide a detailed evaluation of 11 KG embedding methods on link prediction tasks | Laplacian, SVD, GF, HOPE, GraRep, DeepWalk, node2vec, struc2vec, LINE, SDNE, GAE | Quantitative | Compare the performance of 11 different embedding methods on 3 link prediction tasks: drug-disease association, drug-drug interaction, protein-protein interaction and 2 node classification tasks: prediction of protein function and medical term semantic type | AUC | Provide an overview of different types of graph embedding methods, and discuss how they can be used in 3 important biomedical link prediction tasks: DDAs, DDIs and PPIs prediction, and 2 node classification tasks, protein function prediction and medical term semantic type classification<br><br>Compile 7 benchmark datasets for all the above prediction tasks and use them to systematically evaluate 11 representative graph embedding methods selected from different categories<br><br>Develop an easy-to-use Python package with detailed instructions, BioNEV | TRUE | NA |

| Article | Topic | Objective[a] | Approach | | | Evaluation | | | Contributions[a] | Publicly Available | |
|---|---|---|---|---|---|---|---|---|---|---|---|
| | | | Input Data | Overview | Tools | Type | Strategy | Metric(s) | | Code | Interface |
| Relation Prediction Of Co-morbid Diseases Using Knowledge Graph Completion(37) | Disease Comorbidity | Propose a tensor factorization based approach for biological KGs | GO, HPO, DOID, GOA, HPO annotations, STRING, SIDER, Reactome, OMIM, GAD, Medicare claims data | KG: The KG is constructed from the input data sources. Additional edges between diseases and protein-protein interactions were added by clustering<br><br>KG completion: The KG is completed using link prediction | ComplEx | Quantitative | Experiments were performed to verify KG including the prediction of new edges and binary classification of existing triples using the OpenKE framework | Precision, Recall, F1-Score, MRR, Hits@N | Proposed model generates a random vector for entities and relations in the KG and updates it accordingly with the learning of the model by minimizing the loss function<br><br>Deal with anti-symmetric relations by introducing complex vector embeddings and the hermitian dot product in a biological KG | TRUE | NA |
| Sparklis Over Pegase Knowledge Graph: A New Tool For Pharmacovigilance (34) | Pharmacovigilance | Present a novel approach to enhance the way pharmacovigilance specialists perform search and exploration on their data | MedRA, OntoADR, UMLS, SNOMED CT, FAERS | Review existing tools: reviewed over 900 papers from the literature and examined four french pharmacovigilance tools<br><br>KG: Was built from input data sources with some minor modifications to OntoADR and SNOMED CT<br><br>Finally, the Sparklis query builder was extended to better handle clinical terminology hierarchies and accommodate full-text search and multi-selection | Semantic web tools | Both | Two human factors specialists performed and evaluation of Sparklis according to a list of requirements. Then, a cognitive walkthrough of the Spaklis tools was performed | NA | Built a KG that integrates: MedDRA, OntoADR, SMQs, and patient data.<br><br>Conducted a literature review along with a benchmark of existing tools, which led to the detection of several requirements that pharmacovigilance tools have to meet, and which were taken into account in a Sparklis extension.<br><br>Carried out an evaluation of the interface following an ISO ergonomics standard | TRUE | NA |

| Article | Topic | Objective[a] | Approach | | | Evaluation | | | Contributions[a] | Publicly Available | |
|---|---|---|---|---|---|---|---|---|---|---|---|
| | | | Input Data | Overview | Tools | Type | Strategy | Metric(s) | | Code | Interface |
| OWL-NETS: Transforming Owl Representations For Improved Network Inference (38) | Biological Relationships | Propose OWL-NETS, a novel computational method that reversibly abstracts OWL-encoded biomedical knowledge into a network representation tailored for network inference | KaBOB | Created algorithm to remove OWL metadata from a KG subset by: constructing a query graph and representing NETS nodes, identification of NETS edges, creation of network node and edge metadata, and construction of OWL-NETS abstraction network | SPARQL | Both | Three queries were built to examine different biological relationships. For two of the queries, link prediction was performed using 10 different algorithms. Additionally, for all queries, the KG subset pre- and post-removal of OWL-metadata were visualized using Cytoscape<br><br>Validation: A domain expert reviewed the top 20 predicted edges for 2 of the 3 queries | AUC, Top-L precision | Introduce a novel abstraction network methodology that generates semantically rich network representations that are easily consumed by network inference algorithms<br><br>Provided expert-verified evidence from the literature for 50-75% of inferred edges | TRUE | TRUE |

| Article | Topic | Objective[a] | Approach | | | Evaluation | | | Contributions[a] | Publicly Available | |
|---|---|---|---|---|---|---|---|---|---|---|---|
| | | | Input Data | Overview | Tools | Type | Strategy | Metric(s) | | Code | Interface |
| Applying Knowledge-driven Mechanistic Inference To Toxicogenomics (40) | Toxicology | Present an MechSpy, which can be used as a hypothesis generation aid to narrow the scope of mechanistic toxicology analysis | PheKnowLator KG, AOP Wiki, UniProt proteins, Gene Expression data, Manually-curated toxicity mechanisms | KG: Modified PheKnowLator deductively-closed KG by adding human protein-protein interactions and data from the AOP Wiki. KG embeddings were then generated<br><br>Gene expression data: Analyzed microarray and gene expression time series data using traditional bioinformatics workflows<br><br>Mechanisms: Manually-curated mechanisms were embedded, used to generate an enrichment score, and then re-combined with the original KG to create a narrative, and graphical explanation for the 3 most enriched mechanisms | ELK, limma, Robust multiarray average, Cosine similarity | Both | Manually evaluated top scoring mechanism narratives. All assays with at least 1 statistically significant prediction, for the highest chemical dose, for each cell type were eligible for inclusion. For these studies the top-enriched mechanisms were evaluated.<br><br>Validation: Experimental validation was performed for adapin and chlorpromazine | Top-N Precision | Present a framework that holds great potential to aid the hypothesis generation process of mechanistic toxicology<br><br>Combining data from two sources, experimental results and existing knowledge, presents the best of both worlds: this is neither a purely data-driven inference without regards to context, nor a purely semantic knowledge-based exercise of what is plausible<br><br>Given the economic, practical and ethical burden in animal models to elucidate mechanisms of toxicity, MechSpy can also serve as a potential animal testing reduction or replacement tool<br><br>MechSpy has a direct application to both preclinical drug development and pharmacovigilance later on, to study rare side effects on subsets of the population | TRUE | NA |

| Article | Topic | Objective[a] | Approach | | | Evaluation | | | Contributions[a] | Publicly Available | |
|---|---|---|---|---|---|---|---|---|---|---|---|
| | | | Input Data | Overview | Tools | Type | Strategy | Metric(s) | | Code | Interface |
| A Knowledge Graph-based Approach For Exploring The Us Opioid Epidemic (35) | Opioid Abuse | Create the ODKG -- a network of opioid-related drugs, active ingredients, formulations, combinations, and brand names | ATC, RxNorm, EHR Data | KG: The descendants of five base opioid-related classes in the ATC terminology are retrieved: Opioid analgesics, Opioid anesthetics, Opium alkaloids and derivatives, Drugs used in opioid dependence, and Peripheral opioid receptor antagonists and connected to active ingredients in RxNorm using the following edge relations: Ingredients Of, Has Form, Form Of, Part Of, Ingredient Of, Consists Of, Constitutes, Has Trade Name, and Precise Ingredient Of<br><br>EHR Data: Data from more than 400 hospitals was de-identified and normalized | MedEx | Both | A network representation of the KG was built and examined, additionally, the efficacy of MedEx was also evaluated. The KG was also used along with the EHR data to generate summary statistics across the United States | Counts, GIS | The KG is the first KG that captures how opioid drugs relate to each other<br><br>The KG makes it straightforward to translate medications from diverse EHRs into a common set of chemical-dosage features, which subsequently enables a large number of prediction and modeling tasks | TRUE | NA |

| Article | Topic | Objective[a] | Approach | | | Evaluation | | | Contributions[a] | Publicly Available | |
|---|---|---|---|---|---|---|---|---|---|---|---|
| | | | Input Data | Overview | Tools | Type | Strategy | Metric(s) | | Code | Interface |
| Investigating Plausible Reasoning Over Knowledge Graphs For Semantics-based Health Data Analytics (30) | Healthcare Analytics | Propose the SeDan framework that integrates plausible reasoning with expressive, fine-grained biomedical ontologies | DrugBank, DOID, SemMedDB | SeDan framework includes three main modules: Plausible Reasoner to re-write queries based on plausible patterns and ontologies, Knowledge Sources are used to support querying and execute re-written queries, and a User Interface. | RDF, RDFS, OWL, SPARQL, OWL 2 QL, CGLLR | Both | Experiments were run over the framework using queries built from factoid and treatment or diagnosis questions from BioASQ. The results of examining the ability of this system to answer the query were examined by domain experts. | Correctness, Query execution time | The SeDan framework introduces a plausible reasoner that executes a query rewriting algorithm, based on plausible patterns that are both hierarchical and ordered-based and relies on a plausible OWL extension to support (partial) order-based patterns<br><br>SeDan conducts a pattern-driven exploration of the semantic KG to discover hidden associations-compared to graph data mining approaches, these patterns do not need to necessarily occur frequently within the data<br><br>Experiments showed that SeDan expanded the KB by resolving 45% and 11% of unanswered questions about causes and treatments, and 23% overall. Further evaluation shows most of the results are clinically reasonable and verified by a domain expert | NA | TRUE |

| Article | Topic | Objective[a] | Approach | | | Evaluation | | | Contributions[a] | Publicly Available | |
|---|---|---|---|---|---|---|---|---|---|---|---|
| | | | Input Data | Overview | Tools | Type | Strategy | Metric(s) | | Code | Interface |
| Interpretable Graph Convolutional Neural Networks For Inference On Noisy Knowledge Graphs (39) | KG Method Development | Provide a new formulation for GCNNs for link prediction on graph data that addresses common challenges for biomedical KGs | None | Propose a novel method that extends the traditional graph convolutional neural network by adding an attention model where each link has an independent, learnable weight designed to approximate the usefulness of that link | GCNN | Both | Experiments were run to evaluate the novel method using FB15k-237 data as well as a biomedical KG built using full-text PubMed articles, CTD, KEGG, OMIM, BioGRID, Omnipath, and ChEMBL. The biomedical KG was evaluated by investigating the drivers of CF. Finally, the biomedical KG was visualized and manually examined for usefulness | MRR, Hits@10 | Introduces an improvement to graph convolutional neural networks by adding an attention parameter for the network to learn how much to trust an edge during training. As a result, noisier, cheaper data can be effectively leveraged for more accurate predictions<br><br>Facilitates new methods for visualization and interpretation, including ranking the influencers of a node, inferring the greatest drivers of a link prediction, and uncovering errors present in input data sources | NA | NA |
| Structured Reviews For Data And Knowledge Driven Research (45) | Systematic Review KG | Propose structured review articles as KGs focused on specific disease and research questions | Wikidata, Monarch initiative plattform, Molecular signature database, Drosophila RNA-seq data, BioThings, and HMD | KG: Entities and relations were normalized and the KG was constructed using the input data sources<br><br>Hypotheses: Expert-curated questions to be explored in KG<br><br>A Python module was built to integrate and build the KG | Cypher, Neo4j | Qualitative | Provide several visualizations and interpretations of results. Additionally, SemMedDB was used to evaluate the manual curation of disease-based biocuration portion of the workflow | NA | Built the first review article for NGLY1 Deficiency research<br><br>Demonstrated that is now an actionable knowledge resource for the whole community by illustrating how it supports knowledge discovery and dissemination, and it facilitates collaboration between experimental researchers and bioinformaticians | TRUE | TRUE |

| Article | Topic | Objective[a] | Approach | | | Evaluation | | | Contributions[a] | Publicly Available | |
|---|---|---|---|---|---|---|---|---|---|---|---|
| | | | Input Data | Overview | Tools | Type | Strategy | Metric(s) | | Code | Interface |
| GrEDeL A Knowledge Graph Embedding Based Method For Drug Discovery From Biomedical Literatures (57) | Drug Discovery | Propose a biomedical KG embedding-based recurrent neural network method called GrEDeL, which discovers potential drugs for diseases by mining published biomedical literature | MEDLINE, UMLS, TTD | KG: Construct a KG from SemRep and then KG embeddings are constructed<br><br>Model: Apply a RNN with LSTM in order to capture the dependencies between entities of drug-disease associations | SemRep, TransE, LSTM | Both | Drug targets were predicted and existing methods were compared to the proposed approach. Next, a drug rediscovery test was performed<br>Baselines: logistic regression, RF, SVM<br><br>Drug rediscovery<br>Baselines: basic random walk method, NRWRH, TP-NRWRH, Malas's method, Bakalb's method, and SemaTyP<br><br>Six case studies were performed and the run times were determined and compared by embedding length | MRR, Hits@10, Run time | Consider the process of literature-based discovery as a series analysis problem<br><br>Propose a KG based deep learning framework for LBD<br><br>Employs deep learning method combined with KG for drug discovery<br><br>Demonstrate the usefulness of graph embedding-based features for predicting potential drug disease associations | TRUE | NA |
| SemaTyp: A Knowledge Graph Based Literature Mining Method For Drug Discovery (97) | Drug Discovery | Propose a biomedical KG-based drug discovery method called SemaTyP, which discovers candidate drugs for diseases by mining published biomedical literature | PubMed abstracts, UMLS semantic network | KG: Constructed to be a multi-relational graph populated using SemRep data<br><br>Feature Selection: Using explored paths, optimal features are selected | LR, SemKG, SemTyP, SemRep, MetaMap | Quantitative | Using ten-fold cross validation, NRWRH and TP-NRWRH are compared to SemTyP for predicting drug disease associations | Random walk, Precision, Recall, F-score, NRWRH, TP-NRWRH | Introduced a biomedical KG - SemKG - which is constructed by integrating information extracted from PubMed abstracts.<br><br>First method that discovers candidate drugs by using biomedical KG | NA | NA |

| Article | Topic | Objective[a] | Approach | | | Evaluation | | | Contributions[a] | Publicly Available | |
|---|---|---|---|---|---|---|---|---|---|---|---|
| | | | Input Data | Overview | Tools | Type | Strategy | Metric(s) | | Code | Interface |
| Using Predicate Information From A Knowledge Graph To Identify Disease Trajectories (98) | Drug Discovery | Examine the potential added benefit of incorporating data provenance or metadata information into their KG and hypothesized that doing so would result in better performance on machine learning tasks | EKP, UMLS, OMIM, MEDI-HPS, Erasmus Medical Centre | Added provenance information to EKP bu matching identifiers. Then, RFs were trained to predict drug-disease associations by linking paths between drug target proteins and disease proteins | Metab-2-MeSD, RF, caret, pROC, PRROC | Quantitative | Compare the performance of the modified KG to two references sets on a machine learning task designed to predict drug target-disease protein combinations | AUROC, Guney drug-disease combinations | Demonstrated provenance information adds value to drug efficacy screening | NA | NA |
| Using A Knowledge Graph For Hypernymy Detection Between Chinese Symptoms (99) | Diagnostics | Propose a new method which use a KG to detect Chinese symptom hypernym relationship | Medical literature, six healthcare websites | Build a new system to detect hypernym using IS_A relations in SNOMED CT. To do this a symptom component KG is constructed using medical literature to represent atom-symptoms, body parts, and headwords | Bidirectional maximal matching | Quantitative | Compare method to other state of the art methods, adapting them to their corpora<br><br>Other methods compared to are: String Containing, Character Set containing, Feature Vector, Projection Learning, and Simple RNN | Precision, Recall, F1-Score | Proposed method achieved state of the art performance in detecting hypernym between Chinese symptoms | NA | NA |
| Automatic Diagnosis With Efficient Medical Case Searching Based On Evolving Graphs (17) | Clinical Diagnostics | Propose an automatic diagnosis method based on patient records which takes into account the temporal nature of patient records | MIMIC-III, OH | Use evolving graphs to represent how the patient record changes over time. They can efficiently index and compute similarity among these graphs | Unit graph, Evolving graphs, Reverse index, Graph similarity | Both | Used two datasets for evaluation MIMIC-III and OH. There were three components to their evaluation. First was measuring the query speed, second was a measure of accuracy to baseline, finally was to see if their method helped doctor decision making | Average query time, Average accuracy, AUROC, QA | Propose evolving graphs where each graph represents the patient record at each time point. They also introduce several techniques for quickly computing graph similarity | NA | NA |
| Leveraging Distributed Biomedical Knowledge Sources To Discover Novel Uses For Known Drugs (56) | Drug Discovery | Propose a novel system for drug discovery which integrates disparate biomedical knowledge through the use of a KG | Reactome, BioLink, KEGG, UniProtKB, ChEMBL, Pharos, PC2, DOID, OMIM, MONDO, GO, HPO, NCBI Gene, UBERON, Monarch SciGraph, DisGeNet, MyChem.info | Build a KG using 20 different publicly available data sources and representing over 11 distinct types of bio entities and then generated node embeddings which are fed into a RF model to generate novel drug-disease pairs | RF, SVM, LR, node2vec, MyChem API, SemMedDB, The National Drug File, Neo4j | Both | Compare alternative methods ability to predict the probability of drugs treating diseases<br><br>Studies and data from Columbia Open Health data were used as real world examples for validation | F1-Score, AUROC | Integrating many sources of knowledge and using node2vec to train a RF classifier allowed for accurate prediction of drug treating diseases | TRUE | NA |

| Article | Topic | Objective[a] | Approach | | | Evaluation | | | Contributions[a] | Publicly Available | |
|---|---|---|---|---|---|---|---|---|---|---|---|
| | | | Input Data | Overview | Tools | Type | Strategy | Metric(s) | | Code | Interface |
| | | | , miRGate, GeneProf | | | | | | | | |
| Automatic Relationship Verification In Online Medical Knowledge Base: A Large Scale Study In SemMedDB (100) | Automatic Medical Reasoning Verification | Verify medical relationships in SemMedDB | SemMedDB, UMLS | Created a weighted version of SemMedDB, where the edges were semantic types and the weight was derived by determining the number of papers that exist of that type | Naive Bayes, LR, RFt, Decision Tree, kNN, Scikit-Learn | Quantitative | Using several machine learning classifiers where the goal was to determine the accuracy of the edge types | Wilcoxon significance test, Precision, Recall, F1-Score, Accuracy | The proposed features were suitable to train classifiers for evaluating the relations in SemMedDB | NA | NA |
| Gene Ontology Causal Activity Modeling (Go-Cam) Moves Beyond Go Annotations To Structured Descriptions Of Biological Functions And Systems (19) | Biological Relationships | To increase the utility of GO annotations for interpretation of genome-wide experimental data, GO-CAM was developed, a structured framework for linking multiple GO annotations into an integrated model of a biological system. Expect that GO-CAM will enable new applications in pathway and network analysis, as well as improve standard GO annotations for traditional GO-based applications | GO, RO | Formulated a schema to causally link GO molecular activities. Molecular activities are enabled by an active entity (gene product or macromolecular complex) and may take additional entities as input. The activities occur in a location and are parts of biological processes | Noctua | NA | NA | NA | GO-CAMs extend the functionality of the GO by allowing molecular activities to be causally linked. GO-CAMs can be lossily converted back to GO annotations | TRUE | TRUE |

| Article | Topic | Objective[a] | Approach | | | Evaluation | | | Contributions[a] | Publicly Available | |
|---|---|---|---|---|---|---|---|---|---|---|---|
| | | | Input Data | Overview | Tools | Type | Strategy | Metric(s) | | Code | Interface |
| Co-occurrence Graphs For Word Sense Disambiguation In The Biomedical Domain (67) | WSD | Describe an unsupervised method for word sense disambiguation by creating a KG based on co-occurrence which is used to choose between word senses found via dictionary lookup | Abstracts downloaded from PubMed | Use Metamap to annotate articles, then compute a co-occurrence graph. The possible senses of the word are found using dictionary lookup, and the surrounding words in the sentence are annotated using Metamap. The word senses are disambiguated by feeding the context sentence annotations and the possible word senses through the PageRank algorithm | PageRank, MetaMap, Co-occurrence graph | Quantitative | A co-occurrence matrix was built for each corpus and an evaluation was performed to determine the methods ability to correctly disambiguate a set of instances. Baselines includes acronym graph, NLM graph, and a joint graph. Additional baselines include MFS, Metamap, AEC, JDI, MRD, and 2MRD, in addition to several others | Accuracy | Unlike other state-of-the-art techniques, in the proposed method, external resources are not used for the disambiguation step<br><br>Evaluation on two widely used test datasets shows that the reported method obtains consistent results that outperform most of the knowledge-based systems addressing the same problem<br><br>Experiments suggest that the convergence of the method is fast regarding the number of abstracts used for building the graph<br><br>Better results are obtained with less restrictive graphs, since they incorporate to the co-occurrence graph the most useful information about relations between concepts for performing the disambiguation | NA | NA |
| Named Entity Recognition In Traditional Chinese Medicine Clinical Cases Combining BiLSTM-CRF With Knowledge Graph (68) | NER | Propose TCMKG-LSTM-CRF model that utilizes KG information to strengthen the learning ability and recognize rare words | Unspecified textbooks and databases, UMLS | This is a NER task that focuses on recognition of rare words, which is very applicable to TCM | Bi-LSTM, CRF, MetaMap | Quantitative | Manually curated a KG of traditional Chinese medicine from textbooks and other sources<br><br>Implemented an attention mechanism between a hidden layer and KG candidate vectors, that also considered the influence from the previous word | F1-score | Model introduces knowledge attention vector model to implement attention mechanism between hidden vector of neural networks and KG candidate vectors and consider influence from previous word<br><br>Experiment results prove the effectiveness of model | NA | NA |

| Article | Topic | Objective[a] | Approach | | | Evaluation | | | Contributions[a] | Publicly Available | |
|---|---|---|---|---|---|---|---|---|---|---|---|
| | | | Input Data | Overview | Tools | Type | Strategy | Metric(s) | | Code | Interface |
| Barack's Wife Hillary: Using Knowledge Graphs For Fact-aware Language Modeling (69) | Language Model | Introduce the KGLM, a neural language model with mechanisms for selecting and copying facts from a KG that are relevant to the context | WikiText-2 | A language model that incorporates information from the KG to generate more realistic factual knowledge | LSTM, neural-el, Stanford CoreNLP, TransE | Quantitative | Performed NER on wikitext sentences, and liked the entities to their corresponding WikiData concepts<br><br>Created a generative story from all relations to each entity as they are used. Each concept in WikiData has an associated embedding | Perplexity, Fact completion ratio | Proposed KGLM, a neural language model that can access an external source of facts, encoded as a KG, in order to generate text<br><br>Showed that by utilizing this graph, the proposed KGLM is able to generate higher-quality, factually correct text that includes mentions of rare entities and specific tokens like numbers and dates | TRUE | NA |

| Article | Topic | Objective[a] | Approach | | | Evaluation | | | Contributions[a] | Publicly Available | |
|---|---|---|---|---|---|---|---|---|---|---|---|
| | | | Input Data | Overview | Tools | Type | Strategy | Metric(s) | | Code | Interface |
| Multi-task Identification Of Entities, Relations, And Coreference For Scientific Knowledge Graph Construction (65) | Scientific IE | Introduce a multi-task setup of identifying and classifying entities, relations, and coreference clusters in scientific articles | SemEval 2017 Task 10, SemEval 2018 Task 7 | This is a multi-task learning study (NER, RE and coreference resolution) to then summarize each of these detected entities and relations into a KG | Bi-LSTM | Quantitative | Using multiple bidirectional LSTM networks, identify named entities and correferences, as well as the relations between them if applicable<br><br>Apply heuristics to the above entities to map them to a concept in the new KG (or create a new concept). Similarly, add new edges for predicates from the detected relations | F1-score, Precision/pseudo-recall curve | Create a new dataset and develop a multi-task model for identifying entities, relations, and coreference clusters in scientific articles.<br><br>Multi-task setup effectively improves performance across all tasks<br><br>Multi-task model is better at predicting span boundaries and outperforms previous state-of-the-art scientific IE systems on entity and relation extraction, without using any hand engineered features or pipeline processing<br><br>Automatically organize the extracted information from a large collection of scientific articles into a KG<br><br>Analysis shows the importance of coreference links in making a dense, useful graph | TRUE | NA |

| Article | Topic | Objective[a] | Approach | | | Evaluation | | | Contributions[a] | Publicly Available | |
|---|---|---|---|---|---|---|---|---|---|---|---|
| | | | Input Data | Overview | Tools | Type | Strategy | Metric(s) | | Code | Interface |
| Biomedical Term Normalization Of EHRs With UMLS (101) | Term Normalization | Presents a novel prototype for biomedical term normalization of EHR excerpts with the UMLs Metathesaurus | EHR data, UMLS | Uses UMLS to find entity mentions in EHRs. Uses Lucene to perform fast dictionary lookups. The pipeline handles different linguistic aspects in sequence (first expanding acronyms, then spans, finds matches, then disambiguates) | UMLS Metathesaurus | Quantitative | Obtained two English-Spanish parallel corpora and compared annotations produced by Metamap to their proposed method using Cohen's Kappa | Cohen's Kappa | Presented a prototype to perform biomedical term normalization in clinical texts with the UMLS Metathesaurus<br><br>The tool performs abbreviation/acronym expansion and WSD<br><br>As a preliminary evaluation, agreement with MetaMap has been measured in two parallel corpora; the best system has reached moderate agreement with MetaMap<br><br>Presented a web-based user interface for the prototype | NA | TRUE |
| Incorporating Domain Knowledge Into Medical NLI Using Knowledge Graphs (102) | Relation Extraction | Use both word embeddings and node embeddings from a KG to infer relationships such as entailment or contradiction, between a given premise and a hypothesis | UMLS | Given a premise and a hypothesis, this study shows an attempt to classify the pair as entailment, contradiction, or neutral | ESIM, NLTK, MetaMap, DistMult, BioELMo | Both | Use words or phrases from the MedNLI dataset to create a KG from the matching UMLS concepts<br><br>Generate word, node and edge embeddings<br><br>Extract word tokens from the premise and hypothesis using NLTK, link to concepts with MetaMap (and also identify them as biomedical or not), and feed the combination of embeddings of all these to ESIM which performs the classification | Accuracy | Showed that KG embeddings obtained through applying state-of the-art model like DistMult from UMLS could be a promising way towards incorporating domain knowledge leading to improved state-of-the art performance for the medical NLI task<br><br>Further showed that sentiments of medical concepts can contribute to medical NLI task as well | NA | NA |

| Article | Topic | Objective[a] | Approach | | | Evaluation | | | Contributions[a] | Publicly Available | |
|---|---|---|---|---|---|---|---|---|---|---|---|
| | | | Input Data | Overview | Tools | Type | Strategy | Metric(s) | | Code | Interface |
| A New Method For Complex Triplet Extraction Of Biomedical Texts (103) | Relation Extraction | Generate triples from biomedical text to create a KG | NYT Annotated Corpus, WebNLG Dataset, 2010 i2b2/VA Relation Corpus (partial) | This uses a convolution layer and RNN to capture a relation and its components at once. An encoder/decoder model performs the preliminary detection from the sentences | Bi-LSTM, RCNN, NLTK | Quantitative | For each sentence, detect the entities and use their corresponding word embeddings to first predict a relation, then the constituent entities | F1-score | In this model, sentences are encoded by RCNN, which combines the advantages of BiRNN and CNN flexibly, containing more information of sentence<br><br>Experimental results on biomedical dataset and general field dataset show that our method is effective | NA | NA |
| Semantic Relation Extraction Aware Of N-gram Features From Unstructured Biomedical Text (70) | Relation Extraction | Perform NER and relation extraction to generate triples to construct a KG | GENIA corpus, EPI corpus | Use TNG instead of LDA to perform relation extraction over GENIA and EPI. The goal was to overcome LDA's bag-of-word assumption in order to incorporate N-Gram features | LDA, TNG, collapsed Gibbs sampling | Quantitative | Yet another study of "detect entities, detect relations, generate a new triple". | F1-score | Two alternative models, named as Rel-TNG and Type-TNG, are proposed with the help of TNG model and collapsed Gibbs sampling algorithm is utilized for inference<br><br>Extensive experimental results on GENIA and EPI corpora indicate that Rel-TNG and Type-TNG models have similar performance with their unigram counterparts, but Rel-TNG and Type-TNG models outperform Rel-LDA and Type LDA models when prior knowledge is available | NA | NA |
| Long-tail Relation Extraction Via Knowledge Graph Embeddings And Graph Convolutional Networks (66) | Relation Extraction | To improve link prediction on heavily unbalanced datasets as a proxy to RE | NYT dataset | Combine knowledge embeddings with convolutional nets to learn explicit knowledge, and integrate these with the KG using attention. The goal was to leverage existing knowledge to help with relation extraction for low-frequency relations | TransE, GCN | Quantitative | Pre-train node and edge embeddings from the KG. Concatenate the GCN output with the corresponding embeddings, and feed to an attention model | AUPR, Hits@k, Precision@N | Approach provides fine-grained relational knowledge among classes using KG and GCNs, which is quite effective and encoder-agnostic | TRUE | NA |

| Article | Topic | Objective[a] | Approach | | | Evaluation | | | Contributions[a] | Publicly Available | |
|---|---|---|---|---|---|---|---|---|---|---|---|
| | | | Input Data | Overview | Tools | Type | Strategy | Metric(s) | | Code | Interface |
| MedTruth: A Semi-supervised Approach To Discovering Knowledge Condition Information From Multi-source Medical Data (62) | Clinical Enrichment | Propose MedTruth, a method which incorporates source quality into the reliability estimation of the knowledge is acquired | EHRs, Q&A data | Use EHRs as high quality sources of knowledge triple conditionals p(breast_cancer \| female) < p(breast_cancer \| female) to supervise the extraction of knowledge triple conditionals from medical QA. The approach is semi-supervised | MedTruth | Quantitative | Performed several experiments on both real-world datasets and synthetic datasets | Pearson's correlation, Error rate, Confidence | Incorporates knowledge triple conditionals into their knowledge construction method. When pulling knowledge triples from QA sources, each author is treated as a source whose reliability os estimated | NA | NA |
| Towards Smart Healthcare Management Based On Knowledge Graph Technology (104) | Nutrition | Propose a healthy diet KG construction model that promotes the development of healthcare management | Healthcare websites | Performed named entity recognition using CRFs, relation extraction using SVMs, and compute entity relevance using decision trees. Then built the KG from extracted relations | CRF, SVM, Decision trees | Quantitative | Each NLP step is evaluated individually using data collected and annotated with a data schema from three healthcare websites | F1-Score | A model which transforms natural language from healthcare websites into a KG | NA | NA |

| Article | Topic | Objective[a] | Approach | | | Evaluation | | | Contributions[a] | Publicly Available | |
|---|---|---|---|---|---|---|---|---|---|---|---|
| | | | Input Data | Overview | Tools | Type | Strategy | Metric(s) | | Code | Interface |
| NormCo: Deep Disease Normalization For Biomedical Knowledge Base Construction (64) | Disease Normalization | Present NormCo, a deep coherence model which considers the semantics of an entity mention, as well as the topical coherence of the mentions within a single document | MESH, BioASQ, CTD, OMIM, PubMed abstracts | Given a document, the model aims to map mentions of entities to disease concepts. This is accomplished by using entity phrase and coherence models. Models were trained using the input data sources | BiGRU, Transfer learning, NLTK, Word2vec | Quantitative | Disease concept embeddings were created using a subset of BioCreative V Chemical/Disease Relations and the NCBI disease corpus. Baseline methods included mention only, mention and coherence, distantly supervised data, and synthetic data, DNorm, TaggerOne | TF-IDF, Micro-F1-Score, Accuracy, NLCAD, Training time | NormCo models entity mentions using a simple semantic model which composes phrase representations from word embeddings, and treats coherence as a disease concept co-mention sequence using an RNN rather than modeling the joint probability of all concepts in a document<br><br>Experimental results show that NormCo outperforms state-of-the-art baseline methods on two disease normalization corpora in prediction quality and efficiency | TRUE | NA |

[a]Whenever possible, paper objectives and primary results were copied or closely paraphrased from the original manuscripts.

Abbreviations: AOP, Adverse Outcome Pathways; API, Application Programming Interface; ATC, Anatomical Therapeutic Chemical Classification System; AUC, Area Under the Curve; AUPR, Area Under the Precision/Recall curve; AUPRC, Area Under the Precision/Recall Curve; AUROC, Area Under the Receiver Operating Characteristic curve; BH, Benjamini-Hochberg; BiGRU, Bidirectional Gated Recurrent Unit Neural Network; BiLSTM, Bidirectional Long-Short Term Memory; Bio2BEL, Enabling Biological Expression Language; BioASQ, Biomedical Semantic QA; BioGRID, Biological General Repository for Interaction Datasets; BioKEEN, Biological KnowlEdge EmbeddiNgs; CBioMedKG, Chinese biomedical knowledge graph; CDC, Centers for Disease Control; CDM, Common Data Model; CEMR, Chinese Electronic Medical Records; CF, Cystic Fibrosis; CNN, Convolutional Neural Network; ComplEx, Complex Embeddings for Simple Link Prediction; CRF, Conditional Random Field; CTD, Comparative Toxicogenomics Database; DCG, Discounted Cumulative Gain; DDA, Drug-Disease Association; DDI, Drug-Drug Interaction; DDx, Disease Diagnostics; DGDI, Drug-gene-disease Interaction; DM, Dynamic Memory; DMNC, Dual Memory Neural Computer; DOID, Human Disease Ontology; DTI, Drug-Target Interaction; EHR, Electronic Health Record; EKP, Euretos Knowledge Platform; EM, Expectation-Maximization; EPI, Epigenetics and post-translational modification; ERMLP, Entity Recognition Multilayer Perceptron; ESIM, Enhanced Sequential Inference Model; ExPASy, Expert Protein Analysis System; FAERS, Food and Drug Administration Adverse Event Reporting System ; FB15k-237, Freebase-15K-237; GAD, Genetic Association Database; GAE, Graph Autoencoders; GAMENet, Graph Augmented Memory Networks; GCN, Graph Convolutional Network; GCNN, Graph Convolutional Neural Network; GF, Graph Factorization; GIS, Geographic Information System; GO, Gene Ontology; GO-CAM, Gene Ontology Causal Activity Modeling; GOA, Gene Ontology Annotations; GraRep, Graph Representations with Global Structural Information; GrEDeL, Graph Embedding based Deep Learning; HIPPIE, Human Integrated Protein–Protein Interaction rEference; HMD, Human metabolome database; HOPE, High-Order Proximity Embeddings; HPO, Human Phenotype Ontology; HSDN, Human Symptoms-Disease Network; i2b2, Informatics for Integrating Biology & the Bedside; IBM, International Business Machines; ICD10, International Classification of Disease, 10th Revision; IE, Information Extraction; KaBOB, Knowledge Base Of Biomedicine; KEGG, Kyoto Encyclopedia of Genes and Genomes; KG, Knowledge Graph; KGLM, Knowledge Graph Language Model; KnetMiner, Knowledge Network Miner; LBD, Literature Based Discovery; kNN, k Nearest Neighbors; LDA, Latent Dirichlet Allocation; LEAP, LEArn to Prescribe; LINE, Large-Scale Information Network Embedding; LSTM, Long Short-Term Memory; MATADOR, Manually Annotated Target and Drug Online Resource; MB, Memory Bank; MDE, Multi Distance Knowledge Graph Embeddings; MechSpy, Mechanism Spy; MEDI-HPS, MEDI high precision subset; MedRA, Medical Dictionary for Regulatory Activities; MeSH, Medical Subject Headings; MGI, Mouse Genome Informatics; MONDO, Monarch Disease Ontology ; MRR, Mean reciprocal rank; MSigDB, Molecular Signatures Database; NA, Not Applicable; NCBI, National Center for Biotechnology Information; NDCG, Normalized Discounted Cumulative Gain; NDF-RT, National

Drug File - Reference Terminology; NER, Named Entity Recognition; NGLY1, N-glycanase 1; NLCAD, Normalized lowest common ancestor distance; NLI, Natural Language Inference; NLTK, Natural Language Toolkit; NRWRH, Network-based Random Walk with Restart on the Heterogeneous network; NYT, New York Times; OMIM, Online Mendelian Inheritance in Man; OMOP, Observational Medical Outcomes Partnership; OntoADR, Ontology of Adverse Drug Reactions; OWL, Web Ontology Language; OWL-NETS, Web Ontology Language NEtwork Transformation for Statistical learning; PC2, Pathway Commons; PharmGKB, Pharmacogenomics Knowledgebase; PheKnowLator, Phenotype Knowledge Translator Knowledge Graph; PPI, Protein-protein Interactions; ProSNet, Protein function prediction algorithm which efficiently integrates Sequence data with molecular Network data across multiple species; PRROC, Precision-Recall and ROC Curves; PyKEEN, Python KnowlEdge EmbeddiNgs; Q&A, Question and Answer; QL, Query Language; RDB2RDF, Direct Mapping of Relational Data to RDF; RDF, Resource Description Framework; RDFS, Resource Description Framework Schemas; RE, Relation Extraction; RETAIN, REverse Time AttentIoN model; RF, Random Forest; RNA-seq, RNA-sequencing; RNKN, Recursive Neural Knowledge Network ; RNN, Recurrent Neural Network; RO, Relation Ontology; ROBOKOP, Reasoning Over Biomedical Objects linked in Knowledge Oriented Pathways; ROC, Receiver Operating Characteristic; S-MART, Structured Multiple Additive Regression Trees; SDNE, Structural deep network embedding; SE, Structured Embeddings; SeDan, SEmantics-based Data ANalytics; SEM, Simple Event Mode; SemaTyP, Semantic Type Path; SemMedDB, Semantic MEDLINE Database; SGD, Stochastic Gradient Descent; SIDER, Side Effect Resource ; SNOMED CT, Systematized Nomenclature of Medicine -- Clinical Terms; SPARQL, SPARQL Protocol and RDF Query Language; SVD, Singular Value Decomposition; SVM, Support Vector Machine; t-SNE, T-distributed Stochastic Neighbor Embedding; TCM, Traditional Chinese Medicine; TF-IDF, Term Frequency–Inverse Document Frequency; TNG, Topical n-gram model; TP-NRWRH, Two-Pass Random Walk with Restart on the Drug-Disease Heterogenous Network; TransD, Translating on Dynamic Mapping Matrix; TransE, Translations in Embedding Space; TransH, Translating on Hyperplanes; TransR, Translating in Separate Entity Space and Relation Spaces; TTD, Therapeutic Target Database; UBERON, Uber-Anatomy Ontology; UMLS, Unified Medical Language System; UMLS, Unstructured Model; VisAGE, Visualization Assisted by Knowledge Graph Enrichment; WSD, Word-Sense disambiguation.

Supplemental Table 2 Reviewed Articles on Constructing Knowledge Graphs

| Article | Topic | Objective[a] | Approach | | | Evaluation | | | Contributions[a] | Publicly Available | |
|---|---|---|---|---|---|---|---|---|---|---|---|
| | | | Input Data | Overview | Tools | Type | Strategy | Metric(s) | | Code | Interface |
| Liberating Links Between Datasets Using Lightweight Data Publishing: An Example Using Plant Names And The Taxonomic Literature (71) | Biodiversity | Describe the creation of a datasette for a longstanding but mostly unpublished project on linking plant names in The Index of Plant Names Index to the taxonomic literature | IPNI | Preprocessing: Data was downloaded via API, text citations were matched to digital identifiers (i.e. DOIs, Handles, JSTOR links), and finding missing citation information. Then, the data is input into a database, which is implemented in a container | Docker, MySQL | Qualitative | Created several queries to demonstrate available data | NA | Demonstrated the ability to successfully map basic metadata from the IPNI to relevant identifiers in the scientific literature | TRUE | TRUE |
| Ozymandias: A Biodiversity Knowledge Graph (72) | Biodiversity | Explore the feasibility of constructing a "biodiversity KG" for the Australian fauna | Atlas of Living Australia, Australian Faunal Directory, TDWG Life Sciences Identifier vocabulary, TAXREF, Biodiversity Literature Repository, ORCID | A KG was constructed to represent taxa, taxonomic names, publications, journals, and people. The KG was populated using the input data sources and by verifying citation name for all resources using CrossRef | BlazeGraph, Heroku, SPARQL, CrossRef, BioStor OpenURL, CouchDB, Elasticsearch | Qualitative | Provided an overview of the user interface<br><br>Presented high-level statistics to demonstrate the frequency of relevant publications overtime, publication venues, dates of publication by citation dates, and history of species discovery | NA | Created a KG containing representing a taxonomy of Australian fauna and provenance such as data sources and publications<br><br>Provide a web interface for exploring the constructed KG and a SPARQL endpoint | TRUE | TRUE |
| OpenBiodiv-O: Ontology Of The OpenBiodiv Knowledge Management System (73) | Biodiversity | Introduce OpenBiodiv-O, an ontology that fills the gaps between ontologies for biodiversity resources, such as DarwinCore-based ontologies, and semantic publishing ontologies, such as the SPAR Ontologies | FaBiO, DoCO, DwC, Darwin-SW, NOMEN, ENVO, PROTON, TNSS | An ontology was developed by performing domain analysis, identifying important resources and the relationships between them and examination of existing data models and ontologies | RDF | Qualitative | Use case describing how a particularly complicated species was represented in the ontology | NA | Provided a conceptual model of the structure of a biodiversity publication and the development of related taxonomic concepts | TRUE | TRUE |
| Challenges In The Construction Of Knowledge Bases For Human Microbiome-disease Associations (74) | Microbiome | Survey the existing tools and development efforts that have been produced to capture portions of the information needed to construct a KG of all known human microbiome-disease associations and highlight the need for additional innovations in NLP, text mining, taxonomic | NA | Describes the existing KGs for microbiome-disease associations and then provide a detailed overview of how to construct a KG within this domain | NA | None | NA | NA | Four microbiome KGs to capture human microbiome-disease associations have been published, all within the last 2 years<br><br>Identified limitations preventing the development of the more relevant KGs including: tools for microbial entity extraction, annotated corpora of microbial entities, existing catalogs and taxonomies, tools for | NA | NA |

| Article | Topic | Objective[a] | Approach | | | Evaluation | | | Contributions[a] | Publicly Available | |
|---|---|---|---|---|---|---|---|---|---|---|---|
| | | | Input Data | Overview | Tools | Type | Strategy | Metric(s) | | Code | Interface |
| | | representations, and field-wide vocabulary standardization in human microbiome research | | | | | | | disease entity extraction and normalization, and human disease catalogs and taxonomies | | |
| Construction Of The Literature Graph In Semantic Scholar (105) | Scientific Knowledge Management | Describe a deployed scalable system for organizing published scientific literature into a heterogeneous graph to facilitate algorithmic manipulation and discovery | Full-text articles with metadata from publishers, catalogs, and PubMed Central | KG: A KG is constructed using papers, authors, entities, and entity types as nodes and citations, authorship, entity-linking, mention-mention, and entity-entity relations as edges.<br><br>Entity linking: statistical, hybrid, and off-the-shelf methods were explored<br><br>Entity extraction: leverage an approach which does not identify entity types; encode labels using the BILOU scheme | Apache PDFBox, BiLSTM, Viterbi decoding, TagMe, MetaMap Lite, CNN, RNN-LM-LSTM | Quantitative | Leveraged the Sem-Eval-2017 task in order to explore 3 instantiations of different entity extraction models (2 trained on BC5CDR and DHEMDNER and 1 trained on computer science articles from Wikipedia). Then, built a KG from the UMLS and DBpedia. Finally, differences in the entity linking modules were examined through the creation of two datasets are built (1 from combining MeSH and BC5CDR and 1 from combining computer science Wikipedia articles and DBpedia) | F1-Score, Bag of Concepts F1-Score | Discuss the construction of a graph, providing a symbolic representation of the scientific literature<br><br>Describe deployed models for identifying authors, references and entities in the paper text, and provide experimental results to evaluate the performance of each model | TRUE | NA |
| Towards A Knowledge Graph For Science (106) | Scientific Knowledge Management | Propose the vision of a KG for science and present a possible infrastructure for building such a KG | NA | Propose building the science KG by: developing a data model to semantically represent articles which leverages RDF and Linked Data and includes provenance information, creating a scalable storage solution and API, and building a useful user interface | Ontologies, RDF, Neo4j | None | NA | NA | A detailed outline and overview of the steps used to construct a semantic science graph | NA | NA |

| Article | Topic | Objective[a] | Approach | | | Evaluation | | | Contributions[a] | Publicly Available | |
|---|---|---|---|---|---|---|---|---|---|---|---|
| | | | Input Data | Overview | Tools | Type | Strategy | Metric(s) | | Code | Interface |
| Distantly Supervised Biomedical Knowledge Acquisition Via Knowledge Graph Based Attention (107) | Scientific Knowledge Management | Proposed a new end-to-end KGC model | UMLS, MEDLINE | KCG: Combine Prob-TransE and Prob-TransD<br><br>Relation Extraction: Perform sentence representation learning and apply a KG-based attention mechanism to discriminate noisy and informative sentences | ComplEx, CNN, SimplE | Quantitative | RE: distantly supervised RE model is evaluated by comparing triples from MEDLINE to KG triples. Several different models are compared including: RE+KG Prob-TransD, RE+KG Prob-TransE (baselines), RE+KG ComplEx+SimplE, CNN+AVE, and CNN+ATT<br><br>KCG+RE: Link prediction task was performed in order to measure the impact of using RE with KCG | P@N, Precision-Recall curve, MRR | Propose a new end-to-end KGC model, which incorporates word embedding based entity type information into a state-of-the-art KGC model<br><br>Extended model achieves significant and consistent improvements on the biomedical dataset as compared with baseline models | NA | NA |
| The Role Of: A Novel Scientific Knowledge Graph Representation And Construction Model (108) | Scientific Knowledge Management | Propose a novel representation of SciKG, which has three layers. The first layer has concept nodes, attribute nodes, as well as the attaching links from attribute to concept. The second layer represents both fact tuples and condition tuples | MEDLINE abstracts | KG: Built with layers: concept and attribute notes, fact tuples, and conditional statement sentences<br><br>Model: The novel systems combines several powerful NLP methods in order to incorporate all aspects of a traditional workflow, while learning complex dependencies and labeling sequences<br><br>Self-Training: Association rule-based correction, ttag consistency correction and deletion, short sentences only, and deleting incomplete sequences were all strategies employed during the self-training component of the models | BiLSTM-LSTM, Word2vec, Language Model, BERT, POS, Entity recognition, CNN | Both | Proposed approach was evaluated on statistical methods for sequence labeling and the other was OpenIE systems that extract tuples without considering concept attributes or conditions<br><br>Data: Domain experts were asked to manually annotate facts from a random sample of 31 documents<br><br>Baselines: SVM, CRF, AllenNLP Open IE, Stanford Open IE | Precision, Recall, F1-Score | A novel SciKG representation: The new structure represents facts and conditions in scientific statements The SciKG has three layers: statement layer, fact/condition tuple layer and concept/attribute layer<br><br>A novel SciKG construction model: propose a semisupervised multi-input multi-output sequence labeling model for tag prediction and tuple extraction<br><br>Proposed model outperforms baseline methods on a huge literature data | NA | NA |
| Mining Scholarly Publications For Scientific Knowledge Graph Construction (109) | Scientific Knowledge Management | Present a preliminary approach that uses a set of NLP techniques for extracting entities and relationships from research publications and then integrated them in a KG | PubMed abstracts, W2LE, E2E, word embeddings from Microsoft Academic Graph | IE: utilized framework defined by Multi-task identification of entities, relations, and coreference for scientific KG construction.<br>KG: Integrate triples from abstracts | OpenIE, CSO Classifier, WordNet, Word2vec, Clustering, Levenshtein Distance, Wu-Palmer Similarity | Quantitative | The coverage and specificity of the proposed framework was determined when using the extractor framework with the CSO and the extractor framework with OpenIE | Coverage, Specificity | Tackle the challenge of knowledge extraction by employing several state-of-the-art NLP and Text Mining tools<br><br>Describe an approach for integrating entities and relationships generated by these tools<br><br>Analyse an automatically generated KG of 10,425 nodes and 25,655 edges | NA | NA |

| Article | Topic | Objective[a] | Approach | | | Evaluation | | | Contributions[a] | Publicly Available | |
|---|---|---|---|---|---|---|---|---|---|---|---|
| | | | Input Data | Overview | Tools | Type | Strategy | Metric(s) | | Code | Interface |
| Biological Knowledge Graph Construction, Search, And Navigation (110) | Biomedical KG Construction | Illustrate how a KG can be constructed out of the biological RDF databases | dbSNP, UniProt, KEGG, OMIM | KG created from input data sources and was text-indexed to enable live search | Solr | None | NA | NA | Demonstrated how databases can be effectively and efficiently converted into an RDF KG<br><br>Discussed the possibility of inferring new information from existing data, and how to score them based on a reliability measure<br><br>Showed how the KG can be accessed using free-text search | NA | NA |
| Constructing Large Scale Biomedical Knowledge Bases From Scratch With Rapid Annotation Of Interpretable Patterns (81) | Biomedical KG Construction | Present a simple and effective method of extracting new facts with a pre-specified binary relationship type from the biomedical literature, without requiring any training data or hand-crafted rules | DisGeNET, CTD | IE: Assume relationships between entities are binary, candidate for relation extraction if co-occur in the same sentence, 1:many relationships between patterns and inputs, select expressive patterns. Using these tenets information is extracted, sentences are filtered to remove hedging, patterns are ranked, top patterns are annotated by experts, and then new patterns are generated and used to complete KGs | NER, NegEx, Clustering, Maximum Levenshtein Distance | Quantitative | Intrinsic and extrinsic evaluations were performed. The intrinsic evaluation was designed to measure the quality of discovered pairs. The extrinsic evaluation is a KG completion task performed using CTD data and DisGeNET | Recall, Specificity, Precision, F1-Score | Propose a number of methods for extracting patterns from a sentence in which two eligible entities co-occur; different types of patterns have different trade-offs between expressive power and coverage<br><br>Presents patterns in a readable way, enabling faster, more reliable human annotation<br><br>Utilizes these seed pairs to rank newly discovered patterns in terms of their compatibility with the existing data The resulting patterns can be used with or without a human in the loop | NA | NA |
| Constructing Biomedical Domain-specific Knowledge Graph With Minimum Supervision (80) | Biomedical KG Construction | Propose a versatile approach for KG construction with minimum supervision based on unstructured biomedical domain-specific contexts including the steps of entity recognition, unsupervised entity and relation embedding, latent relation generation via clustering, relation refinement and relation assignment to assign | PubMed Articles | KG created by processing text to including performing entity recognition, entity and relation embedding, latent label generation, relation refinement, relation assignment, and relation inference | MinHash, Word2vec, K-means, CNN, Stanford Parser, Jaccard Similarity, Silhouette coefficient | Both | Autism was used as the case study. Evaluation was performed by working with doctors to gather relevant autism queries. Using the queries, PubMed articles were retrieved.The method was evaluated on its ability to return relevant articles. Baseline methods included CNN model with normal max-pooling, piecewise CNN, attention model and ONE<br><br>A qualitative analysis was performed to evaluate the | Hits @10, Hits@20, Hits@50 | Proposed a minimally supervised approach for biomedical KG construction capable of extracting open-ended relations with high precision<br><br>The proposed approach is shown to be accurate and effective for KG construction and the constructed KG is sufficient for further KG completion supported by the experimental results | NA | NA |

| Article | Topic | Objective[a] | Approach | | | Evaluation | | | Contributions[a] | Publicly Available | |
|---|---|---|---|---|---|---|---|---|---|---|---|
| | | | Input Data | Overview | Tools | Type | Strategy | Metric(s) | | Code | Interface |
| | | cluster-level labels | | | | | overall approach. Baselines included SemRep and MinSup | | | | |
| Constructing Biomedical Knowledge Graph Based On SemMedDB And Linked Open Data (82) | Biomedical KG Construction And Evaluation | Propose a novel approach to building a BMKG leveraging the SemMedDB and Health Science LOD | SemMedDB, UMLS, Diseasome, SIDER | KG Evaluation: Checked for inconsistent predications in SemMedDB<br><br>KG: Built a KG by extracting predications from SemMedDB and converting them to RDF. Additional semantic relationships were added to the KG from input data sources | D2R Server | Both | Examined 6 types of predicates: "treats", "affects", "interacts_with", "causes", "disrupts", and "prevents" for inconsistencies<br><br>Mined external data sources to identify new edges | Counts | Detected 462,188 conflicting pairs of semantic triples in SemMedDB<br><br>Discovered over 30 new relationships by linking disorders to genes and drugs | NA | NA |
| Towards FAIRer Biological Knowledge Networks Using A Hybrid Linked Data And Graph Database Approach (83) | Biomedical KG Construction And Evaluation | Present our developments to connect, search and share data about genome-scale knowledge networks | NA | KG: Develop a novel ontology for modeling the representation of life sciences entities | RDF, Neo4j | Quantitative | Describe examples of how BioKNO can be used to model knowledge and includes a description of how data from KnetMiner, WikiPathways, and Reactome could be used | NA | BioKNO has been designed to be a reference model beyond the initial choice of the OWL language, which, FAIR-wise, makes them even more interoperable and reusable | NA | NA |
| Construction Of Traditional Chinese Medicine Knowledge Graph Using Data Mining And Expert Knowledge (76) | Clinical Enrichment | Propose to construct a KG of individual TCM expertise through data mining on a small dataset of EHRs, which leverages both the availability of some small datasets and the analysis from experts by saving their conclusions into KG | EHR | IE: Medical elements and relationships were extracted from an EHR and converted into Dr. Li's medical knowledge<br><br>KG: Schema layer is created from Dr. Li's medical knowledge and knowledge from TCM is combined with EHR data by looking at the similarity between EHR datta and Chineese medicines | Entropy-based Similarity | Both | Two experiments were performed:<br><br>Experts in TCM process medical regulations by removing non-conformity relationships using basic theory and interpret the meaning of relationships between concepts<br><br>Treatment procedures were simulated using the KG and the results were compared to a held-out testing set | Counts | Proposes a method that uses data mining to construct a TCM professional KG with sparse but correct datasets and the help of experts<br><br>Experimental reasoning results confirm the rationality of the constructed KG which also is approved by experts | NA | NA |

| Article | Topic | Objective[a] | Approach | | | Evaluation | | | Contributions[a] | Publicly Available | |
|---|---|---|---|---|---|---|---|---|---|---|---|
| | | | Input Data | Overview | Tools | Type | Strategy | Metric(s) | | Code | Interface |
| Integrating Biomedical Research And Electronic Health Records To Create Knowledge Based Biologically Meaningful Machine Readable Embeddings (75) | Clinical Enrichment | Describe a method for embedding clinical features from EHRs onto SPOKE | SPOKE, EHR data, GWAS catalog | Embeddings for clinical features were created using information from an EHR and then incorporated into SPOKE. To accomplish this, the authors utilize de-identified EHR data, which is filtered to include only those patients diagnosed with at least 1 disease represented in SPOKE | PageRank, Z-scores, Fisher–Yates method | Both | The meaningfulness of the embeddings were examined using different mechanistic causes/influences for different BMIs. This analysis was further extended to examine different genotypic traits and phenotypes/diseases related to different BMIs | Rank<br><br>Probability | By connecting EHR data to SPOKE, real-world context to the network is provided, thus enabling the creation of biologically and medically meaningful “barcodes” (ie, embeddings) for each medical variable that maps onto SPOKE<br><br>Show that these barcodes can be used to recover purposely hidden network relationships, such as Disease–Gene, Disease–Disease, Compound–Gene, and Compound–Compound<br><br>Correct the inference of intentionally deleted edges connecting Side Effects to Anatomy nodes in SPOKE | TRUE | TRUE |
| On Building A Diabetes Centric Knowledge Base Via Mining The Web (111) | Healthcare Management | Propose an approach to constructing a DKB via mining the Web | Vertical data portals, ICD 10 | KG built from data is extracted from portals and stored in a relational database. Then, D2R server is used to convert the data to RDF. Data was cleaned using a voting algorithm and then matched to data in an EHR | RDF, RDFS, OWL, DR2, EM | Both | Eight vertical data portals were selected and processed to build a KG.The accuracy of the triples created from this process were evaluated for accuracy<br><br>Evaluation: Seven people were used to verify the data cleaning and instance matching steps | Accuracy | Introduced an approach to constructing DKB<br><br>Develop a distance based Expectation Maximization algorithm to extract a subset from the overall knowledge base forming the target DKB<br><br>Experiments showed that the data in DKB are rich and of high-quality | NA | NA |
| Personalized Health Knowledge Graph (112) | Healthcare Management | Explain the research challenges to designing the PHKG and provide our vision to address known challenges to constructing these types of KGs | NA | The PHKG architecture is described and includes ontologies, linked open data, relevant patient data, and a mechanism reasoning inference engine | NA | None | NA | NA | A detailed description of a PHKG is proposed and challenges and future work are described | TRUE | NA |

| Article | Topic | Objective[a] | Approach | | | Evaluation | | | Contributions[a] | Publicly Available | |
|---|---|---|---|---|---|---|---|---|---|---|---|
| | | | Input Data | Overview | Tools | Type | Strategy | Metric(s) | | Code | Interface |
| Design And Implementation Of Personal Health Record Systems Based On Knowledge Graph (113) | Healthcare Management | Propose a novel approach to design the KG of medical information and to implement an effective PHR system to support individual healthcare management | EHR | KG represented six pieces of patient data are used to create personal health KG including basic information, diagnoses, visits, laboratory reports, vital signs, and indicators for focus group management. This information is mapped onto KG patterns based on global and local knowledge as well as medical instances created by doctors | OWL, RDF | Qualitative | A case study of one hospital was used to generate 96 classes, 25 object properties, and 48 data properties. The use case was examined in Protege | NA | Proposed an approach which represents patient-centered medical data as a personal health record<br><br>The KG of patients' medical data is designed and implemented with existing related ontologies and the clinical data saved in the EMR system<br><br>Presents a novel knowledge base pattern design and proposes a technical architecture of the PHR system to realize individual healthcare management | NA | NA |
| Intelligent Healthcare Knowledge Resources In Chinese: A Survey (114) | Healthcare Management | Surveys the medical resources needed to construct a medical knowledge base and ways to obtain medical resources | NA | Performs a survey of the medical resources needed to construct a useful medical KG and how to go about collecting relevant data. | NA | None | NA | NA | Surveyed Chinese medical resources in the context of AI, NLP, and database construction and utility and identified major problems in them compared to the resources of other countries<br><br>Text format is not conducive to extracting knowledge<br><br>Medical knowledge coverage is incomplete, low quality, and has varying degrees of accessibility | NA | NA |
| Modelling Online User Behavior For Medical Knowledge Learning (115) | Healthcare Management | Propose a TLLFG, which can learn the descriptions of ailments used in internet searches and match them to the most appropriate formal medical keywords | NMEC, XYWY | Transfer learning using latent factor graph to learn descriptions of ailments and match them to formal medical keywords. Incorporates medical domain knowledge and patient-doctor QA data into a unified latent layer | TLLFG, Loopy Belief Propagation | Both | Used an online medical searching application to verify the model. They also performed multi-class disease prediction.<br><br>The model was evaluated on forecasting performance, micro-analysis, and the ability to detect new correlations between ailments and user-described symptoms. Baselines included: TFG, LFG, logistic regression, and SVM | AUC, average precision and recall, average F1-score, page views | Propose a novel probabilistic model based on, TLLFG<br><br>Experimental results showed that the proposed model substantially outperforms baseline methods<br><br>TLLFG is limited to the extraction of semantic features from medical texts and by the accuracy of labels in the QA data, these limitations may cause significant noise, which will affect the model's performance | NA | NA |

| Article | Topic | Objective[a] | Approach | | | Evaluation | | | Contributions[a] | Publicly Available | |
|---|---|---|---|---|---|---|---|---|---|---|---|
| | | | Input Data | Overview | Tools | Type | Strategy | Metric(s) | | Code | Interface |
| Building Causal Graphs From Medical Literature And Electronic Medical Records (116) | Healthcare Management | Novel approach for automatically constructing causal graphs between medical conditions | MEDLINE, UMLS, EHR data | Medical condition causal graph was constructed from the literature and pruned using medical condition embeddings from an EHR. Then, using text embeddings, potential new neighbors are inferred and used to construct an EHR graph. From the causal text-based graph, medical-condition embeddings, inferred neighbors, and EHR graph, a final causal graph is inferred | SemRep, Co-occurrences, Word2vec, node2vec | Both | Included only medical conditions that occur in more than 5,000 patients. From this data causal graphs for Celiac and Atopic dermatitis were constructed. Evaluators were used to verify all possible causal links for each disease group. SemCause, was used a baseline | PPV | Suggest several methods for constructing causal medical condition graphs<br><br>Apply state-of-the-art natural-language-processing techniques for extracting a small and relevant set of diseases for the causal graph, and further prune this graph using correlations found in EHR data | TRUE | None |
| Mining Disease-symptom Relation From Massive Biomedical Literature And Its Application In Severe Disease Diagnosis (117) | Disease Diagnosis | Present a study on mining disease-symptom relation from massive biomedical literature and constructing biomedical KG from the relation | DOID, SYMP, MEDLINE | IE: Pipeline included preprocessing, search, post-processing, and count<br><br>KG: Built to characterize disease symptoms. The relationship between symptoms and diseases is inferred using conditional probabilities | Whoosh, NegEx, Naive Bayes | Both | 15 patient vignettes were used to evaluate the system | Precision@20 | Present a study o System improves over existing disease-symptom relation mining work include by considering concept hierarchy information from medical entity association discovery<br><br>Implementation based on the constructed KG achieved the best performance compared to all other methods | NA | NA |
| OC-2-KB: Integrating Crowdsourcing Into An Obesity And Cancer Knowledge Base Curation System (78) | Obesity And Cancer | An update to OC-2-KB. It is a software pipeline which automatically extracts semantic triples from PubMed abstracts related to obesity and cancer | OC-2-KB, PubMed titles and abstracts | Preprocessing: Data is processed using standard tools in order to create a collection of sentences. From these sentences, triples are created<br><br>Named entity recognition and Predicate extraction and relation detection is performed, relations are classified and then triples are extracted and stored as RDF | NER, Predicate extraction, Relation detection, LIDF-value | Both | Evaluated crowdsourcing through Amazon Mechanical Turk as a means for validating extracted triples. Two rounds of evaluation were performed<br><br>A small sample of the corpus was randomly sampled and manually annotated by two experts | Time spent on manual evaluation, F1-Score | Extended original corpus from 23 random obesity and cancer related Pubmed abstracts to 82 abstracts from systematic review papers<br><br>Curated a preliminary OCK with these abstracts using OC-2-KB and evaluated the query results against this OCKB<br><br>Refined the underlying machine learning models incorporating the crowdsourcing results | TRUE | NA |

| Article | Topic | Objective[a] | Approach | | | Evaluation | | | Contributions[a] | Publicly Available | |
|---|---|---|---|---|---|---|---|---|---|---|---|
| | | | Input Data | Overview | Tools | Type | Strategy | Metric(s) | | Code | Interface |
| Diagnosis Of COPD Based On A Knowledge Graph And Integrated Model (79) | Chronic Obstructive Pulmonary Disease | Construct a KG of COPD to mine and better understand the relationship between diseases, symptoms, causes, risk factors, drugs, side effects, and more | EHR data | KG: Built to represent diseases, symptoms, causes, risk factors, drugs, side effects and using weighting to indicate the degree of association between symptoms in COPD<br><br>Feature Selection: Built a novel method to identify important features from KG and use as input to hybrid decision model | Correlation information entropy | Both | Data: COPD patient data extracted from an EHR<br><br>Feature selection: k-Means clustering was used to identify important features and the contribution of each feature was recorded. Baselines: MDF-RS, DSA-SVM | Accuracy, Recall, Time complexity, Bidirectional coupling, DSA, SVM, K-means | An integrated model for diagnosing COPD based on a KG is proposed<br><br>The multi-dimensional feature S15 combination extracted by the CMFS-η algorithm has an accuracy of 95.01% and specificity and sensitivity are also good<br><br>Compared with other machine learning algorithms, the DSA-SVM algorithm model also has distinct advantages<br><br>The DSA-SVM algorithm was found to be significantly better than the standard SVM algorithm | NA | NA |
| Aero: An Evidence-based Semantic Web Knowledge Base Of Cancer Behavioral Risk Factors (77) | Cancer | Present a prototype, Aero to (1) better organize and provide evidence based CBRF knowledge extracted from scientific literature (i.e., PubMed), and (2) provide users with access to high-quality scientific knowledge, yet easy to understand answers for their frequently encountered CBRF questions | PubMed Abstracts, UMLS, Relation Ontology, and Time event Ontology | KG: Constructed from identifying articles about smoking, alcohol use, physical activity, and obesity cancer risk factors. Information was extracted from the articles to create tiples, concepts and relationships by building an ontology and connecting to other data sources. Finally, the KG was completed by creating links to other linked data sources<br><br>A user interface was created to navigate the ontology | LogMap, RDFLib, GraphDB | Both | Developed a prototype of the system, created SPARQL queries. The query results were manually evaluated according to whether or not they covered the answers on the NCI risk factor factor sheet | Precision, Recall, 1F-score | Curated a semantic web KB (ie, Aero) to better organize high-quality evidence extracted from scientific literature on the relationships between various behavioral risk factors and cancer<br><br>Created the CBRFO ontology to standardize the terms and relations used across different articles<br><br>Experimented with interactive graph-based visualizations to provide consumers with an easy to understand visual representation of the answers to commonly asked CBRF questions, stimulating their visual thinking | NA | NA |

| Article | Topic | Objective[a] | Approach | | | Evaluation | | | Contributions[a] | Publicly Available | |
|---|---|---|---|---|---|---|---|---|---|---|---|
| | | | Input Data | Overview | Tools | Type | Strategy | Metric(s) | | Code | Interface |
| ALOHA: Developing An Interactive Graph-based Visualization For Dietary Supplement Knowledge Graph Through User-centered Design (118) | Dietary Supplements | Improved a novel interactive visualization platform, ALOHA, for the general public to obtain DS-related information through two user-centered design iterations | iDISK | Data: iDISK is a KG of ontologies, dietary supplement ingredients, diseases known to be treated with dietary supplements, symptoms caused as adverse drug reactions by dietary supplements, and products that contain dietary supplements<br><br>Hypothesis making: Hypotheses were derived from discussions with domain experts and the literature was reviewed<br><br>Needs and Requirements: List derived from usability assessments<br><br>ALOHA components: iDISK data used to populate Neo4j, Flask-based backend and API and Angular front end were built | SUS, RRF | Both | Evaluated usability of ALOHA via SUS and four open-ended questions | Usability | Showed that graph-based interactive visualization is a novel and acceptable approach to end-users who are interested in seeking online health information of various domains<br><br>The authors provide 2 use cases to demonstrate the utility of ALOHA | TRUE | NA |
| Separating Wheat From Chaff: Joining Biomedical Knowledge And Patient Data For Repurposing Medications (119) | Drug Repurposing | Present a system that jointly harnesses large-scale electronic health records data and a concept graph mined from the medical literature to guide drug repurposing—the process of applying known drugs in new ways to treat diseases | Maccabi healthcare EHR, SemMedDB | Data: Identify drugs taken by patients via prescription purchase records. Case and control patients for a specific disease are identified<br><br>KG: Built from SemMedDB | Chi-Square, Propensity-Score-Matching, KS, Bonferroni Correction | Both | Identify patients receiving first-time treatment for a specific disease and collect all prescribed medications. The baseline methods this approach is compared to uses correlation. Patients with hypertension and type II diabetes are used to demonstrate the results<br><br>The gold standard data came from PubMed and medical expert review | Precision@2, Precision@5, Precision@10 | Presented a novel methodology that produces candidates for drug repurposing research by jointly leveraging knowledge from a KG of biological processes that is constructed from PubMed and a large medical records database<br><br>Constructed an interactive system that produces high-quality hypotheses by methodically searching through a large number of candidates and providing biological pathways to support the influence of the candidate medications on conditions | NA | NA |

| Article | Topic | Objective[a] | Approach | | | Evaluation | | | Contributions[a] | Publicly Available | |
|---|---|---|---|---|---|---|---|---|---|---|---|
| | | | Input Data | Overview | Tools | Type | Strategy | Metric(s) | | Code | Interface |
| KGDDS: A System For Drug-drug Similarity Measure In Therapeutic Substitution Based On Knowledge Graph Curation (120) | Drug Substitution | Present a system KGDDS for node-link-based bio-medical KG curation and visualization, aiding Drug-Drug Similarity measure | DOID, IDO, NCBITaxon, HPO, DrugBank, SIDER, PubMed | Developed a new architecture for biomedical KG curation designed to support four key principles: Good Graphic Display, Easy Extensibility, Good Interaction, and Domain specific<br><br>KG: KG built to represent disease, infection site, bacteria, animal, symptom, symptom type, situation, complication, and antibiotics. Data was added from DOID, IDO, NCBITaxon, HPO, and DrugBank<br><br>KGDDS: Pharmacology embeddings are created from SIDER, drug mechanisms were created from NDF-RT and were embedded using LINE, and textual similarity was calculated from DrugBank | Owlready, Force-directed layout, Euclidean norm, Hierarchical edge bundling, IDF | Both | Drug similarity was calculated using a sample of 500,000 papers to evaluate antibiotic-relevant medications. Effectiveness was evaluated by having doctors score the similarity of 1,326 drug pairs. Baseline measures included GADES, Res, Wpath, Hybrids, and MedSim<br><br>A drug substitution case study to identify potential substitutions was performed for cefoperazone | Effectiveness score, Pearson correlation coefficient, Spearman rank correlation | Propose a novel neural network model, which leverages pharmacological and pharmaceutical knowledge from external knowledge bases, so as to broader medical knowledge for similarity measures<br><br>Construct a large-scale antibiotic relevant KG to aid the drug-drug similarity computation, addressing the data skewness and knowledge incompleteness<br><br>Based on experimental results on Drugbank, the method proposed herein achieves better performance than existing methods in drug-drug similarity measures | TRUE | TRUE |
| Supporting Shared Hypothesis Testing In The Biomedical Domain (121) | Hypothesis Generation | Propose a method of assembling biological knowledge about biological processes into hypothesis graphs, a computational framework for hypothesis testing | Cartilage degradation ontology | Assemble axioms from an ontology into a graph. Once assembled the graphs are normalized. Provide methods for using the graphs to perform hypothesis testIng and measure confidence. Confidence is provided by assigning evidence to facts in the graph | Protege | Both | Validated proposed method with domain experts from MultiScale Human project and provided a cartilage degradation use case | Confidence | Proposed and evaluated a framework for producing and using hypothesis graphs | TRUE | TRUE |

[a]Whenever possible, paper objectives and primary results were copied or closely paraphrased from the original manuscripts.

Abbreviations: ATT, Attention; AVE, Average Attention; BILOU, Beginning, Inside, Last, Outside, and Unit; BiLSTM, Bidirectional Long-Short Term Memory; BMI, Body Mass Index; CNN, Convolutional Neural Network; ComplEx, Complex Embeddings for Simple Link Prediction; COPD, Chronic Obstructive Pulmonary Disease; CSO, Computer Science Ontology; CTD, Comparative Toxicogenomics Database; Darwin-SW, Darwin Semantic Web; dbSNP, Single Nucleotide Polymorphism Database; DOI, Digital Object Identifier; DSA, Direct Search Simulated Annealing; DwC, Darwin Core; EHR, Electronic Health Record; EM, Expectation-Maximization; ENVO, Environment Ontology; FaBiO, Functional Requirements for Bibliographic Records-Aligned Bibliographic Ontology; GADES, A Graph-based Semantic Similarity Measure; GWAS, Genome-Wide Association Study; ICD10, International Classification of Disease, 10th Revision; IDF, Inverse Document Frequency; iDISK, integrated DIetary Supplement Knowledge base; IDO, Infectious Disease Ontology; IPNI, International Plant Names Index; KCG, Knowledge Graph Convolutional Networks; KEGG, Kyoto Encyclopedia of Genes and Genomes; KG, Knowledge Graph; KS, Kolmogorov-Smirnov; LFG, Latent Factor Graph; LIDF, Linguistic patterns, IDF, and C-value information; LINE, Large-Scale Information Network Embedding; LM, Language Model; LSTM, Long Short-Term Memory; MRR, Mean reciprocal rank; NA, Not Applicable; NCBITaxon, National Center for Biotechnology Information taxonomy database; NER, Named Entity Recognition; NLP, Natural Language Processing; NMEC, National Medical Examination; NOMEN, Nomenclatural Ontology; OC-2-KB, Obesity and Cancer to Knowledge Base; OCKB, Obesity and Cancer Knowledge Base; OMIM, Online Mendelian Inheritance in Man; OpenIE, Open Information Extraction; OWL, Web Ontology Language; POS, Part-Of-Speech; PPV, Positive Predictive Value; PROTON, Proton Ontology; RDF, Resource Description Framework; RDFS, Resource Description Framework Schemas; RE, Relation Extraction; RNN, Recurrent Neural Network; RRF, Rich Release Format; SemMedDB, Semantic MEDLINE Database; SIDER, Side Effect Resource ; SimplE, Simple Embedding; SPOKE, Scalable Precision Medicine Oriented Knowledge Engine; SUS, System Usability Scale; SVM, Support Vector Machine; SYMP, Symptom Ontology; TCM, Traditional Chinese Medicine; TDWG, Biodiversity Information Standards; TFG, Transfer Factor Graph ; TLLFG, Transfer Learning using

Latent Factor Graph ; TNSS, Taxonomic Nomenclatural Status Terms; TransD, Translating on Dynamic Mapping Matrix; TransE, Translations in Embedding Space; UMLS, Unified Medical Language System; URL, Universal Resource Locator.

Supplemental Table 3 Currently available biomedical data science knowledge graphs

| Name | Primitives | Domain | Backend | Last Updated | Construction Method | Endpoint | Availability |
|---|---|---|---|---|---|---|---|
| Bio2RDF (122) | Ontology concepts (URIs) | Biomedical | Virtuoso | 09/25/14 | Semantic integration of OBOs | SPARQL | Download at http://download.bio2rdf.org/#/ Tools at https://github.com/bio2rdf |
| BioGrakn (123) | Ontology concepts (URIs) | Biomedical | Grakn.ai | 09/27/19 | Semantic integration of OBOs | Graql | Download/install from https://github.com/graknlabs/biograkn |
| DisGeNET (124) | Ontology concepts (URIs) | Biomedical | Unspecified | 01/19 | Semantic integration of OBOs | SPARQL | Browse at http://disgenet.org/search Download at http://disgenet.org/downloads |
| HetioNet (125) | Ontology concepts (URIs) | Biomedical | Unspecified | 07/01/19 | Manual Curation | SPARQL, REST, Noctua | Browse and download at https://geneontology.cloud/home |
| KaBOB (46) | Ontology concepts (URIs) | Biomedical | Neo4J | 07/08/19 | Semantic integration of OBOs | Neo4J | Download at https://github.com/hetio/hetionet Tools at https://het.io/software/ |
| NGLY1 Deficiency (45) | Ontology concepts (URIs) | Biomedical | Allegrograph | 06/23/19 | Semantic integration of OBOs | SPARQL | Download and tools at https://github.com/UCDenver-ccp/kabob |
| Ozymandias (72) | Ontology concepts (URIs) | NGLY1 Deficiency | Neo4J | 08/08/19 | Semantic integration of OBOs | Neo4J | Download at https://github.com/SuLab/ngly1-graph Browse at http://ngly1graph.org/browser/ |
| PheKnowLator (61) | Ontology concepts (URIs) | Biodiversity | Blazegraph | 2019 | Linked data from ALA and ALD using CrossRef | SPARQL | Browse at https://ozymandias-demo.herokuapp.com |
| ROBOKOP (63) | Ontology concepts (URIs) | Human Biomedical | None | 09/25/19 | Semantic integration of OBOs | NetworkX, RDF | Download at https://github.com/callahantiff/PheKnowLator/wiki |
| Sparklis (34) | Ontology concepts (URIs) | Biomedical | GreenT, Neo4J | 09/20/19 | Semantic integration of OBOs | Neo4J, JSON API | Download at http://robokopkg.renci.org/browser/ Tools at https://github.com/NCATS-Gamma |
| Bio2RDF (122) | Ontology concepts (URIs) | Pharmacovigilance | Unspecified | 01/19 | Semantic integration of OBOs | SPARQL | Browse at http://www.irisa.fr/LIS/ferre/sparklis/ |

Abbreviations: ALD, Australian Faunal Directory; ALA, Atlas of Living Australia; NGLY1, N-glycanase 1; OBO, Open Biomedical Ontology; URI, Universal Resource Identifier.